\documentclass{article} 
\usepackage{iclr2026_conference,times}
\usepackage{amsmath}
\usepackage{amssymb}
\usepackage{amsthm}
\usepackage{mathtools}
\usepackage{nicefrac}
\usepackage{amsfonts}
\usepackage{stmaryrd}
\SetSymbolFont{stmry}{bold}{U}{stmry}{m}{n}
\usepackage{bbm}
\usepackage{scalerel}

\usepackage{graphicx}
\usepackage{booktabs}      
\usepackage{multirow}      
\usepackage{diagbox}       
\usepackage{makecell}      
\usepackage{colortbl}      
\usepackage{caption}       
\usepackage{subcaption}    
\usepackage{arydshln}      
\usepackage{wrapfig}
\usepackage{capt-of}
\usepackage{tabularx}
\usepackage[export]{adjustbox}
\usepackage{siunitx} 

\usepackage{listings}      
\usepackage{algorithm}     
\usepackage{algorithmic}   

\usepackage[utf8]{inputenc} 
\usepackage[T1]{fontenc}    
\usepackage{microtype}     
\usepackage{enumitem}      
\usepackage{rotating}      
\usepackage{csquotes}      
\usepackage{epigraph}      
\usepackage{pifont}        
\usepackage{varwidth}
\usepackage{xspace}
\usepackage{minitoc}
\usepackage{rotating}
\usepackage{siunitx}

\usepackage{url}
\usepackage{hyperref} 
\usepackage[capitalize,noabbrev]{cleveref} 

\usepackage{tikz}
\usetikzlibrary{tikzmark}

\usepackage{amsmath,amsfonts,bm}

\def\eqref#1{equation~\ref{#1}}

\def\1{\bm{1}}

\def\mR{{\bm{R}}}

\DeclareMathAlphabet{\mathsfit}{\encodingdefault}{\sfdefault}{m}{sl}
\SetMathAlphabet{\mathsfit}{bold}{\encodingdefault}{\sfdefault}{bx}{n}

\newcommand{\fnet}{FloydNet\xspace}
\newcommand{\kfnet}[1]{#1-FloydNet\xspace}

\newtheorem{theorem}{Theorem}
\newtheorem{definition}{Definition}

\newtheorem{proposition}{Proposition}
\newtheorem{corollary}{Corollary}
\newtheorem{lemma}{Lemma}
\newtheorem{remark}{Remark}

\def\1{\bm{1}} 

 \def\mR{{\bm{R}}}

\title{\fnet: Pivotal Attention for Relation Composition}

\author{
Jingcheng Yu$^{1}$\thanks{Equal contribution. Code: \url{https://github.com/ocx-lab/FloydNet}}\ \ \quad Mingliang Zeng$^{1*}$\hspace{2pt}\quad Qiwei Ye$^{1*}$ \\
$^1$Beijing Academy of Artificial Intelligence \\
\texttt{\{jcyu, mlzeng, qwye\}@baai.ac.cn}
}

\iclrfinalcopy 
\begin{document}
\maketitle

\doparttoc 
\faketableofcontents

\vspace{-5pt}


\begin{abstract}
Learning algorithmic computation often requires explicit relational intermediate states, yet many graph processors maintain their primary states on individual entities. We introduce \fnet and \textbf{Pivotal Attention} (PA), which maintain ordered pair states and update a target relation $(i,k)$ by attending over candidates formed from $(i,j)$ and $(j,k)$ for every pivot $j$. Motivated by the pair-and-pivot structure of Floyd--Warshall, PA learns relation composition and pivot weighting in parallel rather than executing its ordered min-plus recurrence. The \kfnet{k} framework extends this operation to ordered $k$-tuples, with Self-Attention and PA as its $k=1$ and $k=2$ cases at the attention-operation level. Under atomic tuple initialization and invariant readout, we show that \kfnet{k} is no more graph-discriminative than k-FWL; on BREC, each evaluated variant matches the success set of its corresponding WL reference. \fnet further achieves 96.64\% mean accuracy under the reported CLRS-30 protocol and a 99.8\% optimality rate with 10 samples on held-out non-metric TSP instances.
\end{abstract}

\section{Introduction}
Neural Algorithmic Reasoning asks whether neural networks can learn representations of algorithmic computation that transfer beyond the instances seen during training \citep{velivckovic2021neural,velivckovic2022clrs}. A central challenge is the choice of intermediate state: a processor must represent not only the input entities, but also the subproblems and partial results created during computation.

Most graph processors keep their primary states on individual nodes. Local message passing propagates these states along graph edges, while Graph Transformers allow global interactions among them \citep{gilmer2017neural,xu2019powerful,dwivedi2020generalization,ying2021transformers}. Both are effective designs, but global communication alone does not make relations explicit computational objects. Many algorithms are naturally expressed through pair- or tuple-indexed states: shortest-path algorithms maintain pairwise distances, geometric systems compose frame transformations, and combinatorial solvers make pairwise decisions. Encoding these states only through node representations leaves the relevant relations implicit.

Explicit relation states also require an update rule that reflects how they interact. In many algorithms, a relation from $i$ to $j$ and a relation from $j$ to $k$ jointly inform the relation from $i$ to $k$. Treating all pair states as unrelated tokens does not expose this shared-pivot structure. This motivates an operator whose candidates are relations that can be composed through an intermediate entity.

We introduce \textbf{Pivotal Attention} (PA). \fnet maintains an ordered pair state $\mR_{ik}$. For each pivot $j$, PA forms a candidate from $\mR_{ij}$ and $\mR_{jk}$, then uses $\mR_{ik}$ to weight and aggregate these candidates. Self-Attention attends over entity states; PA attends over candidate relation compositions. Pair-state and triangle-indexed architectures provide important precedents \citep{jumper2021highly,abramson2024accurate,bergen2021edgetransformer}; PA differs in making both composable relations part of each candidate key and value.

The abstraction is captured by \kfnet{k}. Its states are indexed by ordered $k$-tuples, and a common pivot generates one candidate by replacing each tuple coordinate in turn. At the level of state space and attention operation, $k=1$ gives entity-state Self-Attention, $k=2$ gives pair-state Pivotal Attention, and $k>2$ represents higher-order relations. This unifies their attention operations, not the complete architectures.

Our design is inspired by dynamic programming (DP), which organizes computation through recurrences over reusable subproblem states \citep{bellman1966dynamic}. Floyd--Warshall and its earlier closure lineage provide a canonical example of all-pairs states composed through intermediate vertices \citep{roy1959transitivite,floyd1962algorithm,warshall1962theorem}:
\begin{equation}
D^{(k)}_{ij}=\min\!\left\{D^{(k-1)}_{ij},\;D^{(k-1)}_{ik}+D^{(k-1)}_{kj}\right\}.
\label{eq:floyd_warshall_recurrence}
\end{equation}
Here stage $k$ admits vertex $k$ as an additional intermediate. \fnet uses this pair-and-pivot pattern but not the ordered min-plus recurrence: PA learns the composition and pivot weights and evaluates all candidates in parallel.

The common-pivot construction also mirrors k-FWL. Under atomic tuple initialization and invariant readout, \kfnet{k} is upper-bounded by k-FWL; on BREC, the evaluated variants match their WL references' success sets under RPC. We further evaluate algorithmic computation, pair-structured optimization, and graph prediction.

Our primary contributions are:
\begin{enumerate}[topsep=0pt,leftmargin=15pt,itemsep=-2pt]
    \item We introduce Pivotal Attention, in which each target pair relation attends over pivot-indexed compositions of aligned incoming and outgoing relations.
    \item We express PA as the $k=2$ case of a common-pivot ordered-tuple attention framework that includes entity-state Self-Attention at $k=1$, and provide a fused pairwise implementation that avoids materializing cubic pivot-message tensors.
    \item We prove a k-FWL upper bound under explicit initialization and readout assumptions and evaluate the relation-first inductive bias through BREC, CLRS-30, TSP, and graph prediction benchmarks.
\end{enumerate}

\section{The \fnet Framework}
\label{sec:floydnetwork_architecture}

\subsection{Relational State Spaces for Attention}
Attention refines a target by comparing it with candidate representations. Standard Self-Attention (SA) operates on entity states $\mathbf H_i$, using projected entity keys and values. FloydNet instead maintains an ordered pair state $\mR_{ik}\in\mathbb R^{d_r}$ and refines this relation from candidates composed through intermediate entities.

\subsection{Pivotal Attention: Learning Relations among Relations}
\label{sec:pivotal_attention}
For a target relation $\mR_{ik}$, every entity $j$ aligns two composable relations, $\mR_{ij}$ and $\mR_{jk}$. Their shared endpoint makes $j$ a \emph{pivot}: the first relation ends where the second begins. Pivotal Attention (PA) asks how the relation composed through each pivot should refine the target relation.

Let there be $h$ heads of dimension $d_h$, with $d_r=h d_h$. For head $a\in[h]$, define
\begin{align}
\mathbf q^a_{ik}&=W_Q^a\mR_{ik},\nonumber\\
\boldsymbol\kappa^a_{ijk}&=W_{K,L}^a\mR_{ij}+W_{K,R}^a\mR_{jk},\nonumber\\
\boldsymbol\nu^a_{ijk}&=W_{V,L}^a\mR_{ij}+W_{V,R}^a\mR_{jk}.
\label{eq:pivotal_qkv}
\end{align}
The role-specific projections distinguish the incoming relation $(i,j)$ from the outgoing relation $(j,k)$. The composed key $\boldsymbol\kappa^a_{ijk}$ represents how the two relations jointly match the target, while the composed value $\boldsymbol\nu^a_{ijk}$ represents the information that their composition contributes. PA then normalizes over pivots:
\begin{align}
\alpha^a_{ijk}&=
\frac{\exp\!\left(\langle\mathbf q^a_{ik},\boldsymbol\kappa^a_{ijk}\rangle/\sqrt{d_h}\right)}
{\sum_{p\in V}\exp\!\left(\langle\mathbf q^a_{ik},\boldsymbol\kappa^a_{ipk}\rangle/\sqrt{d_h}\right)},
&
\mathbf o^a_{ik}&=\sum_{j\in V}\alpha^a_{ijk}\boldsymbol\nu^a_{ijk},
\label{eq:pivotal_attn}\\
\operatorname{PA}(\mR)_{ik}&=W_O[\mathbf o^1_{ik}\|\cdots\|\mathbf o^h_{ik}].
\nonumber
\end{align}
Thus SA learns entity-to-entity interactions, whereas PA learns interactions between a target relation and pivot-aligned compositions of other relations. Both use query--key compatibility and value aggregation; they differ in the order of the represented object and in how each candidate message is constructed. The main model uses the additive composition in \cref{eq:pivotal_qkv}. Other composition operators, including elementwise multiplication, are discussed in \cref{sec:appendix_full_version}.

This construction retains the central inductive bias of Floyd--Warshall \citep{floyd1962algorithm,warshall1962theorem}---composing pair states through an intermediate vertex---without executing its fixed min-plus recurrence or ordered pivot schedule. PA learns both the relation composition and the relevance of each pivot, and evaluates all pivots in parallel.

\begin{figure*}[t]
\begin{minipage}{0.45\textwidth}
    \includegraphics[width=1.0\textwidth]{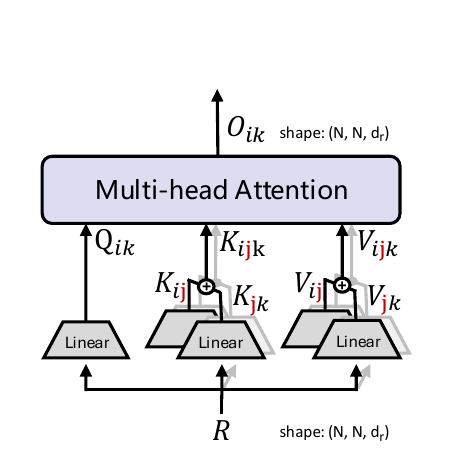}
\end{minipage}
\hfill
\begin{minipage}{0.55\textwidth}
\centering
\begin{lstlisting}[language=python, mathescape, breaklines=true, linewidth=0.98\linewidth]
def PivotAttnCore(q_ik, k_ij, k_jk, v_ij, v_jk):
    # Inputs: [batch, n, n, heads, head_dim]
    k_ijk = (k_ij[:, :, :, None, :, :] +
             k_jk[:, None, :, :, :, :])
    v_ijk = (v_ij[:, :, :, None, :, :] +
             v_jk[:, None, :, :, :, :])
    q_ixk = q_ik[:, :, None, :, :, :]

    a_ijk = torch.sum(q_ixk * k_ijk, dim=-1)
    a_ijk = a_ijk / (q_ik.shape[-1] ** 0.5)
    w_ijk = torch.softmax(a_ijk, dim=-3) # pivot j
    o_ijk = v_ijk * w_ijk[..., None]
    return torch.sum(o_ijk, dim=-4) # pivot j
\end{lstlisting}
\end{minipage}
\caption{\textbf{Pivotal Attention refines a relation through pivot-specific candidates.} To update $(i,k)$, each pivot $j$ provides a candidate built jointly from $(i,j)$ and $(j,k)$. The target state $\mR_{ik}$ supplies the query used to weight and aggregate these candidates.}
\label{fig:method}
\end{figure*}

\subsection{The Unified \kfnet{k} Framework}
\label{sec:general_fnet}
The same operation extends to relations of any order. For $k\geq1$, let
$\mathbf v=(v_1,\ldots,v_k)\in V^k$ be an ordered tuple, with repeated vertices allowed, and let $\mR_{\mathbf v}$ denote its state. For coordinate $r\in[k]$ and pivot $p\in V$, define
\[
\mathbf v[r\leftarrow p]=(v_1,\ldots,v_{r-1},p,v_{r+1},\ldots,v_k).
\]
A common pivot produces the $k$ coordinate-replacement states
$\big(\mR_{\mathbf v[1\leftarrow p]},\ldots,\mR_{\mathbf v[k\leftarrow p]}\big)$.
For head $a$, \kfnet{k} constructs the additive pivot message
\begin{align}
\mathbf q^a_{\mathbf v}&=W_Q^a\mR_{\mathbf v},\nonumber\\
\boldsymbol\kappa^a_{\mathbf v,p}&=
\sum_{r=1}^{k}W_{K,r}^a\mR_{\mathbf v[r\leftarrow p]},
&
\boldsymbol\nu^a_{\mathbf v,p}&=
\sum_{r=1}^{k}W_{V,r}^a\mR_{\mathbf v[r\leftarrow p]},
\label{eq:k_pivotal_messages}\\
\alpha^a_{\mathbf v,p}&=
\frac{\exp\!\left(\langle\mathbf q^a_{\mathbf v},\boldsymbol\kappa^a_{\mathbf v,p}\rangle/\sqrt{d_h}\right)}
{\sum_{u\in V}\exp\!\left(\langle\mathbf q^a_{\mathbf v},\boldsymbol\kappa^a_{\mathbf v,u}\rangle/\sqrt{d_h}\right)},
&
\mathbf o^a_{\mathbf v}&=\sum_{p\in V}\alpha^a_{\mathbf v,p}\boldsymbol\nu^a_{\mathbf v,p}.
\label{eq:k_pivotal_attn}
\end{align}
Coordinate-specific maps preserve tuple roles. For $k=1$, the replacement state is simply $\mR_p$, recovering node-set SA at the operation level; for $k=2$, the replacements $(p,k)$ and $(i,p)$ recover PA after assigning its right and left roles; and $k>2$ represents explicit higher-order relations. This is a state-space and operation-level unification, not an identity between complete architectures. Increasing $k$ enlarges the available relational state space; its formal k-FWL placement is summarized below and proved in \cref{sec:appendix_proofs}.

\subsection{Instantiating FloydNet on Graphs}
\paragraph{Inputs and initialization.}
Let $G=(V,E,\mathbf X,\mathbf E,\mathbf G)$ be a finite attributed graph, where $N=|V|$, $\mathbf X_i\in\mathbb R^{d_n}$ is a node feature, $\mathbf E_{ij}\in\mathbb R^{d_e}$ is the feature of ordered pair $(i,j)$, and $\mathbf G\in\mathbb R^{d_g}$ is optional. We treat $\mathbf E$ as a total pairwise input: nonedges, self-pairs, and observed edges have distinct fixed types whenever relevant. The pairwise model maintains $\mR^{(\ell)}\in\mathbb R^{N\times N\times d_r}$ and initializes
\begin{equation}
\mR^{(0)}_{ij}=\operatorname{Init}\!\left(
\mathbf G,\mathbf X_i,\mathbf X_j,\mathbf E_{ij},\mathbf 1\{i=j\}
\right),
\label{eq:init_rel}
\end{equation}
where $\operatorname{Init}$ is shared over pairs and implemented by an MLP. Inputs already encoded by $\mathbf E_{ij}$ may be omitted.

\paragraph{FloydBlock.}
A FloydBlock applies PA inside a pre-normalized residual block \citep{vaswani2017attention}:
\begin{align}
\widehat{\mR}&=\operatorname{Norm}(\mR^{(\ell-1)}),\nonumber\\
\widetilde{\mR}&=\mR^{(\ell-1)}+\operatorname{PA}(\widehat{\mR}),
\label{eq:floydblock_attn}\\
\mR^{(\ell)}&=\widetilde{\mR}+\operatorname{FFN}(\operatorname{Norm}(\widetilde{\mR})).
\label{eq:floydblock_ffn}
\end{align}
All maps are shared over tuple indices. Norm may be LayerNorm, RMSNorm \citep{zhang2019root}, or query--key normalization as used by \citet{dehghani2023scaling}.

\paragraph{Task readout.}
For node- and graph-level tasks, we may augment $V$ by a distinguished SuperNode $s$, writing $V^+=V\sqcup\{s\}$. It has a unique learnable feature and fixed learnable virtual-edge types, and permutations of original vertices fix $s$. Shared decoders produce
\begin{equation}
\widehat y_i=D_{\mathrm{node}}(\mR^{(L)}_{is}),\qquad
\widehat y_{ij}=D_{\mathrm{pair}}(\mR^{(L)}_{ij}),\qquad
\widehat y_G=D_{\mathrm{graph}}(\mR^{(L)}_{ss}).
\label{eq:task_readouts}
\end{equation}

\begin{figure*}[t]
\centering
\includegraphics[width=\textwidth]{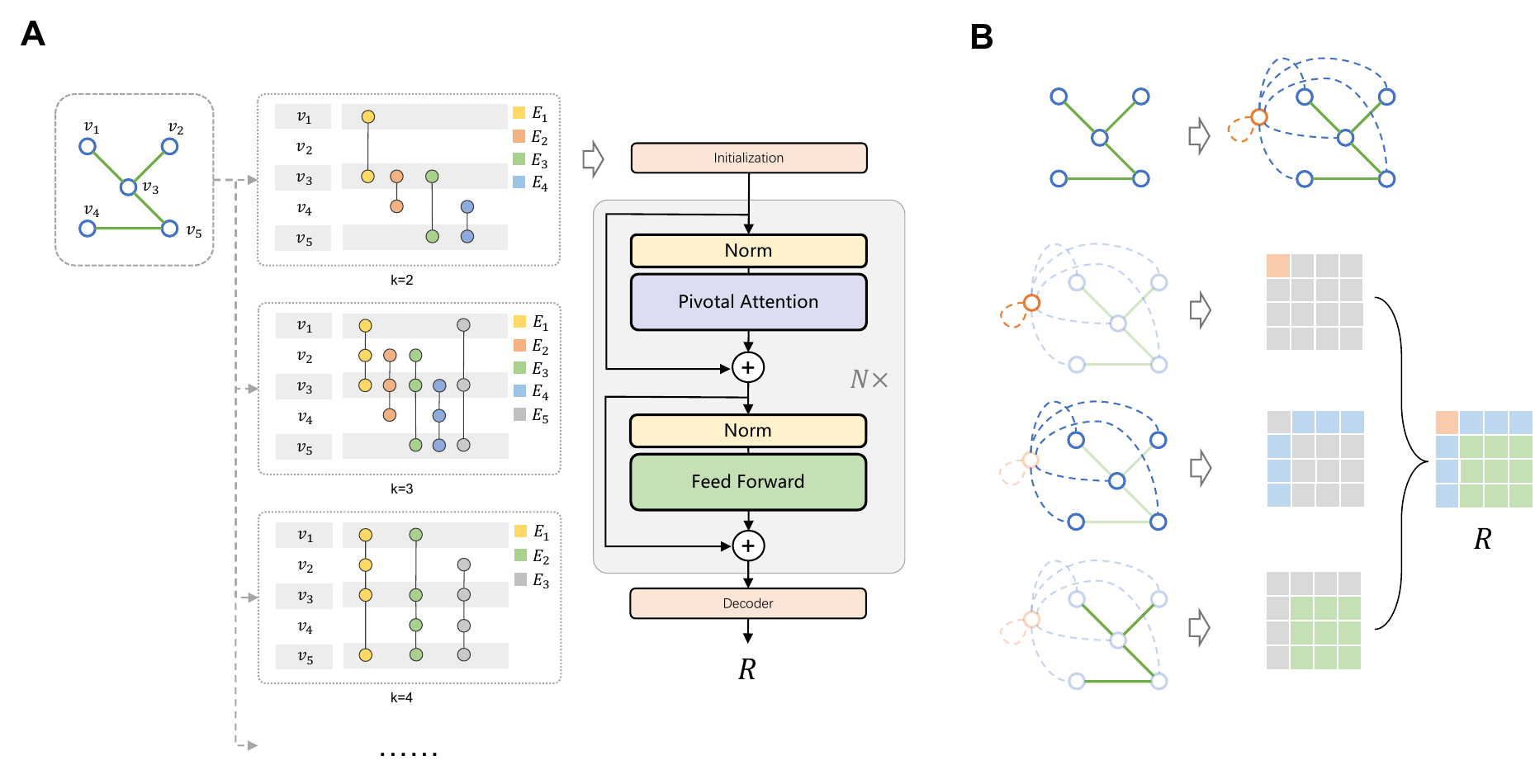}
\caption{\textbf{A unified hierarchy of relational state spaces.} \textbf{(A)} \kfnet{k} lifts attention from entity states ($k=1$) to pair relations ($k=2$) and higher-order ordered-tuple relations ($k>2$). \textbf{(B)} FloydNet instantiates the pairwise case with stacked FloydBlocks and optional SuperNode readouts.}
\label{fig:overview}
\end{figure*}

\subsection{Equivariance and Complexity}
For a permutation $\pi$ of $V$, let $\pi^+$ extend it to $V^+$ with $\pi^+(s)=s$, and define
$(\pi^+\!\cdot\!\mR)_{ik}=\mR_{(\pi^+)^{-1}(i),(\pi^+)^{-1}(k)}$.
\begin{proposition}[Permutation equivariance]
\label{prop:equivariance}
If input features and masks are relabeled with the graph and all tuple maps are shared, then
\[
\operatorname{FloydBlock}(\pi^+\!\cdot\!\mR)=
\pi^+\!\cdot\!\operatorname{FloydBlock}(\mR).
\]
Consequently, node and pair predictions are equivariant, and the graph prediction from $\mR_{ss}$ is invariant.
\end{proposition}
The result follows by reindexing the pivot sum through $p\mapsto\pi(p)$; shared pointwise maps, normalization, and residual connections preserve equivariance. Index-based positional encodings are unnecessary, although equivariant structural features may be used.

For dense tuple states, fixed $k$ and $h$, and FFN width $\Theta(d_r)$, one block takes
\begin{equation}
\mathcal O(N^{k+1}d_r+N^k d_r^2)
\label{eq:k_floyd_complexity}
\end{equation}
time. A naive implementation materializes $\mathcal O(N^{k+1}d_r)$ pivot messages. The fused $k=2$ kernel avoids this materialization and uses $\mathcal O(N^2d_r)$ forward intermediate storage, apart from implementation-dependent training workspace.

\section{k-FWL Placement of \kfnet{k}}
\label{sec:theoretical_analysis}

The common-pivot organization of \kfnet{k} closely mirrors the neighborhood construction of folklore k-FWL, providing a natural formal coordinate system for the architecture. We use the convention in which each ordered $k$-tuple is refined from the multiset, over a common pivot, of the ordered vector of its $k$ coordinate replacements; under this convention, k-FWL has the distinguishing power of ordinary $(k+1)$-WL \citep{cai1992optimal,maron2019provably}.

\begin{theorem}[k-FWL upper bound]
\label{thm:fwl_upper_bound}
Consider \kfnet{k} on finite attributed graphs. Assume that (i) each initial tuple state is a shared function only of its attributed atomic type, including any graph-level attribute supplied to initialization, (ii) all tuple maps are shared over tuple indices, (iii) pivot aggregation is permutation invariant, and (iv) the graph readout is permutation invariant and uses no information beyond the final tuple states. If two ordered tuples have the same k-FWL color after $\ell$ rounds, then their \kfnet{k} states after $\ell$ layers are equal. Consequently, \kfnet{k} cannot distinguish any graph pair that k-FWL fails to distinguish.
\end{theorem}
\begin{proof}[Proof sketch]
The claim follows by induction on $\ell$. It holds at initialization by assumption (i). Equality of the k-FWL colors at round $\ell+1$ implies equality of the previous colors and equality of the multisets of ordered coordinate-replacement color vectors. By the induction hypothesis, the corresponding neural states agree. Shared projections and the symmetric softmax over pivots therefore produce equal attention outputs; residual, normalization, and pointwise FFN maps preserve equality. The invariant readout gives the graph-level conclusion. See \cref{sec:appendix_proofs} for details.
\end{proof}

\begin{corollary}[Pairwise upper bound]
\label{cor:pairwise_upper_bound}
Under the assumptions of \cref{thm:fwl_upper_bound}, \fnet ($k=2$) is at most as graph-distinguishing as 2-FWL, and hence as ordinary 3-WL under the convention above.
\end{corollary}

The theorem formally places the architecture in the WL hierarchy. On BREC, the evaluated variants distinguish exactly the same benchmark pairs as their corresponding WL references under the RPC protocol, linking this placement to realized behavior on that finite benchmark. The formal definitions, full proof, and scope are given in \cref{sec:appendix_proofs}.

\begin{proposition}[Value-composition depth]
\label{prop:dependency_depth}
Fix a target pair and a pivot choice at every layer, and recursively expand only the two value-side pair states composed at each chosen pivot. After $L$ FloydBlocks, this binary value-composition tree has at most $2^L$ initial pair-state leaves.
\end{proposition}
\begin{proof}[Proof sketch]
At $L=0$ the tree consists of one initial pair state. Each value-side composition joins two trees from the preceding layer, so the maximum number of leaves doubles at every layer.
\end{proof}
This proposition describes an individual value-composition tree, not the full dependency graph: attention weights additionally depend on the target query and keys, and the output aggregates over every pivot. Because the implemented model initializes every ordered pair and uses all pivots, it has global computational access from its first layer. Only with an explicit adjacency mask and edge-supported initialization can a selected composition tree be interpreted as concatenating graph walks of length at most $2^L$.

\section{Experiments}
\label{sec:experiments}

We evaluate structural discrimination, algorithmic generalization, TSP optimization, and transfer to graph benchmarks.

Our code, data generation scripts, and exact hyperparameters are available at \url{https://github.com/ocx-lab/FloydNet}. Where applicable (e.g., CLRS), we use the official evaluation harness without hints at test time unless explicitly stated. For TSP, train and test sets have disjoint size ranges and seeds; we report the full benchmark and a filtered subset that reduces multiple-optimum cases.

\subsection{Relation-State Discrimination and Pair-Level Concordance}
We first test whether explicit pair and tuple states support structural counting and graph-pair discrimination.

\paragraph{Quantitative Expressiveness on Homomorphism Counting Tasks.}
We use the synthetic counting tasks of \citet{zhang2024beyond}. These tasks provide an empirical quantitative probe beyond standard 1-WL-limited message passing. As shown in \cref{tab:exp_homo,tab:exp_subgraph}, \fnet obtains substantially lower Mean Absolute Error than the displayed MPNN baseline across the reported graph-, node-, and edge-level tasks.

\paragraph{Qualitative Expressiveness on Isomorphism Testing.}
The BREC benchmark \citep{wang2024empirical} provides 400 pairs of non-isomorphic graphs specifically designed to be indistinguishable by the 1-WL test; a model's accuracy reflects how many pairs it can correctly tell apart. As shown in \Cref{tab:brec_summary}, under its Reliable Paired Comparisons (RPC) protocol, \fnet distinguishes exactly the same graph pairs as the 3-WL reference. This pair-level concordance is stronger than aggregate-score agreement and matches the 3-WL reference success set on BREC, without implying equivalence on arbitrary graphs (\Cref{sec:theoretical_analysis}).

\paragraph{Expressiveness of \kfnet{k}.}
We further evaluate \kfnet{k} on BREC. As summarized in \Cref{tab:brec_summary} and detailed in \Cref{sec:appendix_full_result_brec}, \kfnet{3} and \kfnet{4} achieve accuracies of 95.0\% and 99.8\%, respectively. Each variant distinguishes exactly the same BREC pairs as its corresponding WL reference, extending this pair-level empirical concordance across tuple orders (\Cref{sec:theoretical_analysis}).

\begin{table*}[t]
\centering
\begin{minipage}[t]{0.62\textwidth}
    \centering
\caption{MAE on the evaluated homomorphism-counting tasks. \fnet obtains near-zero MAE on the reported tasks.}
    \label{tab:exp_homo}
    \small
    \setlength{\tabcolsep}{1.5pt}
    \resizebox{\linewidth}{!}{
        \begin{tabular}{c|ccc|cc|ccc}
        \toprule
        \multirow{2}{*}{\backslashbox{Model}{Task}} & \multicolumn{3}{c|}{Graph-level} & \multicolumn{2}{c|}{Node-level}& \multicolumn{3}{c}{Edge-level}\\
        & \adjustbox{valign=c}{\includegraphics[height=0.05\linewidth]{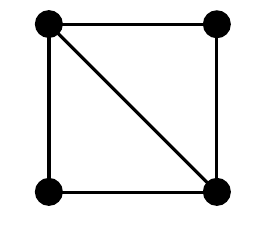}} & \adjustbox{valign=c}{\includegraphics[height=0.05\linewidth]{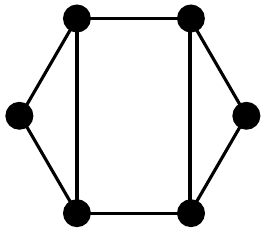}} & \adjustbox{valign=c}{\includegraphics[height=0.05\linewidth]{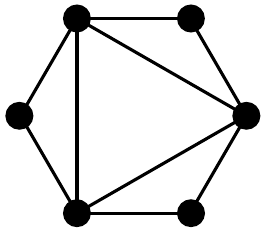}} & \adjustbox{valign=c}{\includegraphics[height=0.05\linewidth]{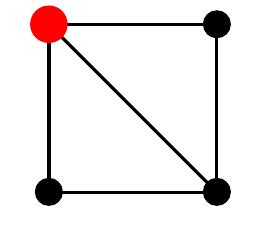}} & \adjustbox{valign=c}{\includegraphics[height=0.05\linewidth]{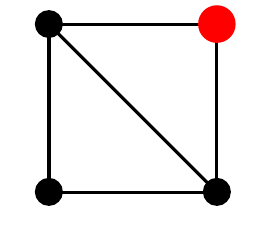}} & \adjustbox{valign=c}{\includegraphics[height=0.05\linewidth]{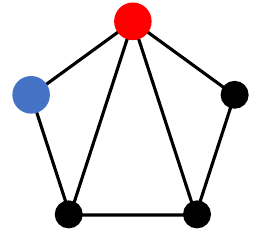}} & \adjustbox{valign=c}{\includegraphics[height=0.05\linewidth]{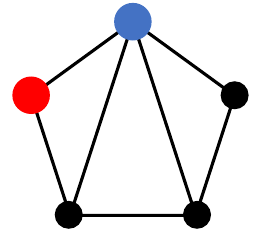}} & \adjustbox{valign=c}{\includegraphics[height=0.05\linewidth]{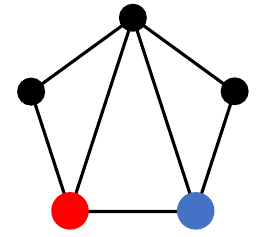}}\\ \midrule
        MPNN (GIN) & \cellcolor{red!20} .300 & \cellcolor{red!20} .233 & \cellcolor{red!20} .254 & \cellcolor{red!20} .505 & \cellcolor{red!20} .478 & - & - & - \\
        Subgraph GNN & \cellcolor{orange!20} .011 & \cellcolor{orange!20} .015 & \cellcolor{orange!20} .012 & \cellcolor{green!20} .004 & \cellcolor{orange!20} .058 & \cellcolor{green!20} .003 & \cellcolor{orange!20} .058 & \cellcolor{orange!20} .048 \\
        Local 2-GNN & \cellcolor{green!20} .008 & \cellcolor{green!20} .008 & \cellcolor{green!20} .010 & \cellcolor{green!20} .003 & \cellcolor{green!20} .004 & \cellcolor{green!20} .005 & \cellcolor{green!20} .006 & \cellcolor{green!20} .008 \\
        Local 2-FGNN & \cellcolor{green!20} .003 & \cellcolor{green!20} .005 & \cellcolor{green!20} .004 & \cellcolor{green!20} .005 & \cellcolor{green!20} .005 & \cellcolor{green!20} .007 & \cellcolor{green!20} .007 & \cellcolor{green!20} .008 \\ \midrule
        \textbf{\fnet} & \cellcolor{cyan!20} .000& \cellcolor{cyan!20} .001& \cellcolor{cyan!20} .001& \cellcolor{cyan!20} .000 & \cellcolor{cyan!20} .000 & \cellcolor{cyan!20} .000 & \cellcolor{cyan!20} .000 & \cellcolor{cyan!20} .000 \\
        \bottomrule
        \end{tabular}
    }   
\end{minipage}
\hfill
\begin{minipage}[t]{0.31\textwidth}
    \centering
    \caption{Accuracy on BREC. Each \kfnet{k} variant distinguishes the same individual pairs as its displayed WL reference; }
    \label{tab:brec_summary}
    \small
    \setlength{\tabcolsep}{4pt}
    \resizebox{\linewidth}{!}{
    \begin{tabular}{lc}
    \toprule
    Model / Heuristic & Accuracy (\%) \\
    \midrule
    MPNN (1-WL equiv.) & \textless 20\% \\
    Graphormer & 19.8\% \\
    PPGT&  58.5\%\\
    KP-GNN & 68.8\% \\
    \midrule
    3-WL(2-FWL) Test & \textbf{67.5\%} \\
    \textbf{\fnet} & \cellcolor{green!20}\textbf{67.5\%} \\
    \textbf{\kfnet{3}} & \cellcolor{cyan!20}\textbf{95.0\%} \\
    \textbf{\kfnet{4}} & \cellcolor{cyan!20}\textbf{99.8\%} \\
    \bottomrule
    \end{tabular}
    }
\end{minipage}
\end{table*}

\begin{table*}[t]
\small
\centering
\caption{MAE on the evaluated (chordal) cycle-counting tasks. \fnet obtains near-zero MAE on the reported tasks.}
\label{tab:exp_subgraph}
\setlength{\tabcolsep}{1.5pt}
\makebox[\textwidth]{\resizebox{\textwidth}{!}{
\begin{tabular}{c|cccccc|cccccc|cccccc}
\toprule
\multirow{2}{*}{\backslashbox{Model}{Task}} & \multicolumn{6}{c|}{Graph-level} & \multicolumn{6}{c|}{Node-level} & \multicolumn{6}{c}{Edge-level} \\
& \adjustbox{valign=c}{\includegraphics[height=0.03\textwidth]{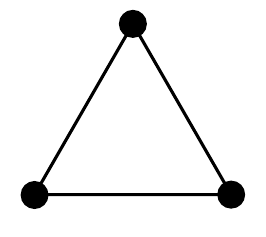}} & \adjustbox{valign=c}{\includegraphics[height=0.03\textwidth]{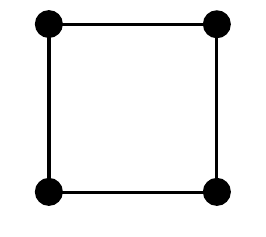}} & \adjustbox{valign=c}{\includegraphics[height=0.03\textwidth]{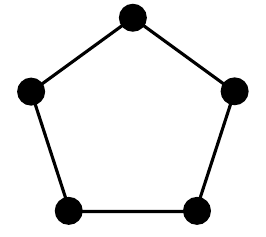}} & \adjustbox{valign=c}{\includegraphics[height=0.03\textwidth]{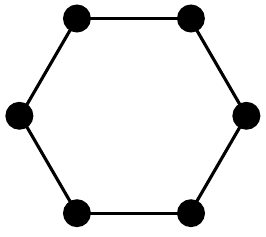}} & \adjustbox{valign=c}{\includegraphics[height=0.03\textwidth]{figure/bwl-figure/exp_chordal_4-cycle.pdf}} & \adjustbox{valign=c}{\includegraphics[height=0.03\textwidth]{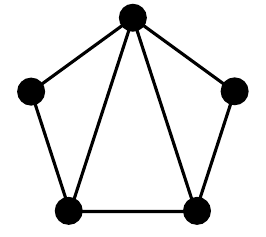}} & \adjustbox{valign=c}{\includegraphics[height=0.03\textwidth]{figure/bwl-figure/exp_3-cycle.pdf}} & \adjustbox{valign=c}{\includegraphics[height=0.03\textwidth]{figure/bwl-figure/exp_4-cycle.pdf}} & \adjustbox{valign=c}{\includegraphics[height=0.03\textwidth]{figure/bwl-figure/exp_5-cycle.pdf}} & \adjustbox{valign=c}{\includegraphics[height=0.03\textwidth]{figure/bwl-figure/exp_6-cycle.pdf}} & \adjustbox{valign=c}{\includegraphics[height=0.03\textwidth]{figure/bwl-figure/exp_chordal_4-cycle.pdf}} & \adjustbox{valign=c}{\includegraphics[height=0.03\textwidth]{figure/bwl-figure/exp_chordal_5-cycle.pdf}} & \adjustbox{valign=c}{\includegraphics[height=0.03\textwidth]{figure/bwl-figure/exp_3-cycle.pdf}} & \adjustbox{valign=c}{\includegraphics[height=0.03\textwidth]{figure/bwl-figure/exp_4-cycle.pdf}} & \adjustbox{valign=c}{\includegraphics[height=0.03\textwidth]{figure/bwl-figure/exp_5-cycle.pdf}} & \adjustbox{valign=c}{\includegraphics[height=0.03\textwidth]{figure/bwl-figure/exp_6-cycle.pdf}} & \adjustbox{valign=c}{\includegraphics[height=0.03\textwidth]{figure/bwl-figure/exp_chordal_4-cycle.pdf}} & \adjustbox{valign=c}{\includegraphics[height=0.03\textwidth]{figure/bwl-figure/exp_chordal_5-cycle.pdf}} \\ \midrule
MPNN & \cellcolor{red!20} $.358$ & \cellcolor{red!20} $.208$ & \cellcolor{red!20} $.188$ & \cellcolor{red!20} $.146$ & \cellcolor{red!20} $.261$ & \cellcolor{red!20} $.205$ & \cellcolor{red!20} $.600$ & \cellcolor{red!20} $.413$ & \cellcolor{red!20} $.300$ & \cellcolor{red!20} $.207$ & \cellcolor{red!20} $.318$ & \cellcolor{red!20} $.237$ & - & - & - & - & - & - \\
Subgraph GNN & \cellcolor{orange!20} $.010$ & \cellcolor{orange!20} $.020$ & \cellcolor{orange!20} $.024$ & \cellcolor{green!20} $.046$ & \cellcolor{orange!20} $.007$ & \cellcolor{green!20} $.027$ & \cellcolor{green!20} $.003$ & \cellcolor{green!20} $.005$ & \cellcolor{orange!20} $.092$ & \cellcolor{orange!20} $.082$ & \cellcolor{orange!20} $.050$ & \cellcolor{orange!20} $.073$ & \cellcolor{green!20} $.001$ & \cellcolor{green!20} $.003$ & \cellcolor{orange!20} $.090$ & \cellcolor{orange!20} $.096$ & \cellcolor{green!20} $.038$ & \cellcolor{orange!20} $.065$ \\
Local 2-GNN & \cellcolor{green!20} $.008$ & \cellcolor{green!20} $.011$ & \cellcolor{green!20} $.017$ & \cellcolor{green!20} $.034$ & \cellcolor{green!20} $.007$ & \cellcolor{green!20} $.016$ & \cellcolor{green!20} $.002$ & \cellcolor{green!20} $.005$ & \cellcolor{green!20} $.010$ & \cellcolor{green!20} $.023$ & \cellcolor{green!20} $.004$ & \cellcolor{green!20} $.015$ & \cellcolor{green!20} $.001$ & \cellcolor{green!20} $.005$ & \cellcolor{green!20} $.010$ & \cellcolor{green!20} $.019$ & \cellcolor{green!20} $.005$ & \cellcolor{green!20} $.014$ \\
Local 2-FGNN & \cellcolor{green!20} $.003$ & \cellcolor{green!20} $.004$ & \cellcolor{green!20} $.010$ & \cellcolor{green!20} $.020$ & \cellcolor{green!20} $.003$ & \cellcolor{green!20} $.010$ & \cellcolor{green!20} $.004$ & \cellcolor{green!20} $.006$ & \cellcolor{green!20} $.012$ & \cellcolor{green!20} $.021$ & \cellcolor{green!20} $.004$ & \cellcolor{green!20} $.014$ & \cellcolor{green!20} $.003$ & \cellcolor{green!20} $.006$ & \cellcolor{green!20} $.012$ & \cellcolor{green!20} $.022$ & \cellcolor{green!20} $.005$ & \cellcolor{green!20} $.012$ \\ \midrule
\textbf{\fnet} & \cellcolor{cyan!20} .000 & \cellcolor{cyan!20} .000& \cellcolor{cyan!20} .001& \cellcolor{cyan!20} .001& \cellcolor{cyan!20} .000& \cellcolor{cyan!20} .001& \cellcolor{cyan!20} .000 & \cellcolor{cyan!20} .000 & \cellcolor{cyan!20} .000 & \cellcolor{cyan!20} .000 & \cellcolor{cyan!20} .000 & \cellcolor{cyan!20} .000 & \cellcolor{cyan!20} .000 & \cellcolor{cyan!20} .000 & \cellcolor{cyan!20} .000 & \cellcolor{cyan!20} .000 & \cellcolor{cyan!20} .000 & \cellcolor{cyan!20} .000 \\
\bottomrule
\end{tabular}
}}
\end{table*}

\subsection{Learned Relation Composition for Algorithmic Reasoning}
We next test whether pivot-mediated relation composition supports algorithmic computation and OOD size generalization.

\begin{figure}[h]
    \centering
    \begin{subfigure}[b]{0.4\textwidth}
        \centering
        \includegraphics[width=\linewidth]{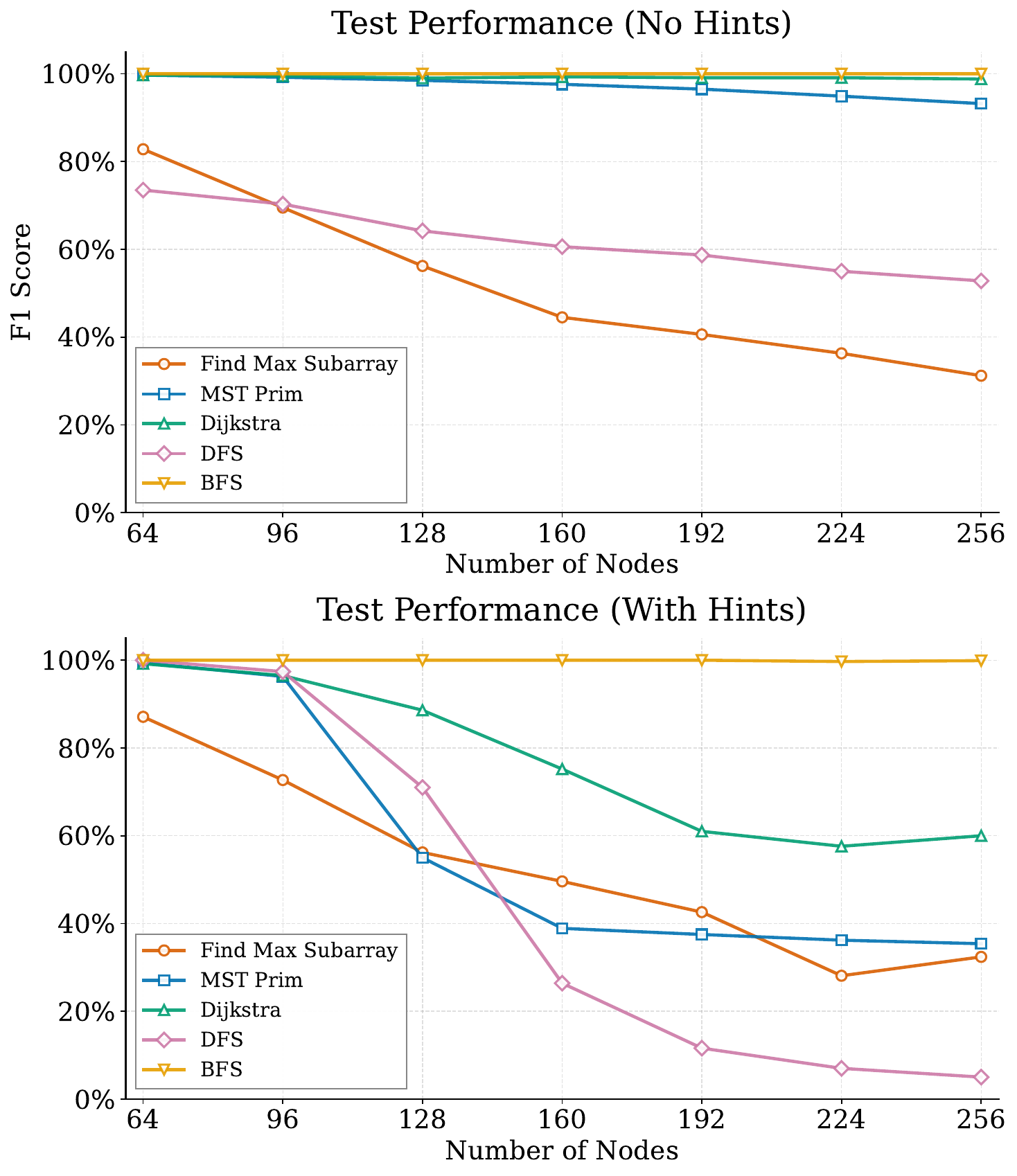}
        \caption{\textbf{OOD Scalability Analysis ($n \to 256$).}}
        \label{fig:probing_a}
    \end{subfigure}
    \hspace{1cm}
    \begin{subfigure}[b]{0.35\textwidth}
        \centering
        \includegraphics[width=\linewidth]{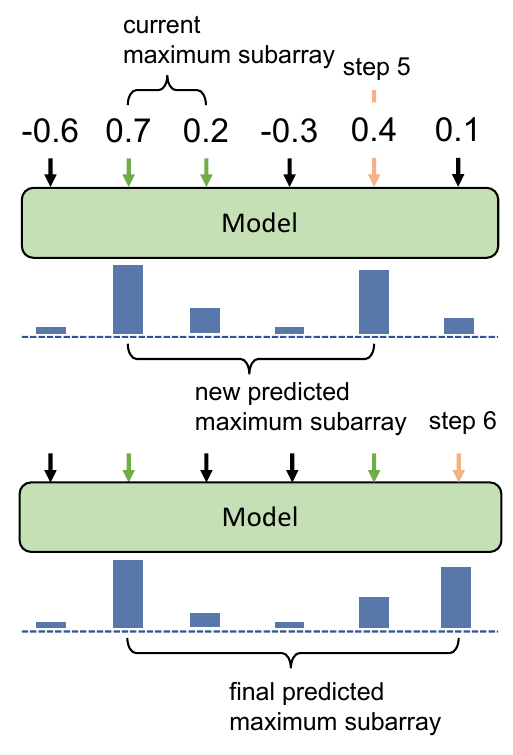}
        \caption{\textbf{Algorithmic Alignment.}}
        \label{fig:probing_b}
    \end{subfigure}
    \caption{\textbf{Probing the Scalability and Reasoning Logic of \fnet on CLRS.} \textbf{(a)} We extend our OOD evaluation to problem sizes up to $n=256$, within the broader extrapolation range studied by \citet{minder2023salsa} but using our own evaluation sizes and protocol. The results indicate that models trained \textit{without hints} (upper) generally exhibit superior length extrapolation capabilities compared to those trained \textit{with hints} (lower). \textbf{(b)} Visualization of the internal reasoning process for the Maximum Subarray problem (Kadane's algorithm). The model iteratively predicts the new maximum subarray based on the current maximum subarray and the next number  corresponding to the current step. The observed alignment provides evidence consistent with learning a state-updating computation, but does not by itself establish faithful execution of the reference algorithm.}
    \label{fig:clrs_analysis}
\end{figure}

\paragraph{The CLRS Benchmark.} The CLRS-30 benchmark \citep{velivckovic2022clrs} is a standard suite for assessing algorithmic reasoning. We train one \fnet model per CLRS-30 task. Evaluation is performed on out-of-distribution instances with larger problem sizes, without providing any ground-truth hints to the model at test time.
We evaluate on the full benchmark and report aggregated performance across eight categories in \Cref{tab:clrs_summary_new}. Under this protocol, \fnet has the highest reported aggregate performance and exceeds $95\%$ on most algorithms, including Strings, graph algorithms such as Floyd--Warshall, and Sorting. In these configurations, removing hints improves most tasks. 
To test length scalability, we extend our OOD evaluation to problem sizes up to $n=256$, within the broader extrapolation range studied by SALSA-CLRS \citep{minder2023salsa} but using our own evaluation sizes and protocol. As visualized in \Cref{fig:clrs_analysis}(a), the no-hint model performs better at these sizes. Full per-algorithm results and detailed extrapolation scores are available in Appendix \ref{sec:appendix_full_result_clrs}.

\paragraph{Probing Algorithmic Reasoning.}
To examine whether \fnet represents intermediate algorithmic structure, we probed its layer-wise states on dynamic programming tasks. We use \textit{Find Max. Subarray (Kadane's algorithm)} as a case study, where the reference solution maintains and updates an optimal subarray during a traversal. As visualized in \Cref{fig:clrs_analysis}b, alignment with the reference intermediate trajectory provides evidence consistent with learning a state-updating computation, rather than relying solely on input--output correlations; it does not by itself establish faithful execution of the reference algorithm.

\begin{table}[h!]
\centering
\small
\caption{Average test accuracy (\%) on the CLRS-30 benchmark, aggregated by algorithm category. $N$ denotes the number of algorithms in each group. All models are trained on problem sizes of $n=16$ and tested on larger instances of $n=64$.}
\label{tab:clrs_summary_new}
\setlength{\tabcolsep}{1pt} 
\resizebox{\columnwidth}{!}{
\begin{tabular}{@{}llccccccccc@{}}
\toprule
\textbf{Class} & \textbf{Model} & \makecell{Sort \\ (N=4)} & \makecell{Search \\ (N=3)} & \makecell{D\&C \\ (N=1)} & \makecell{Greedy \\ (N=2)} & \makecell{DP \\ (N=3)} & \makecell{Graph \\ (N=12)} & \makecell{String \\ (N=2)} & \makecell{Geometry \\ (N=3)} & \makecell{Total \\ (N=30)} \\
\midrule
\multirow{3}{*}{\textbf{GNN}}& \textbf{Triplet-GMPNN} \citep{ibarz2022generalist} & 75.64 & 58.83 & 76.36 & \cellcolor{orange!30}91.22 & 81.99 & \cellcolor{orange!30}86.42 & 49.09 & \cellcolor{orange!30}94.09 & 80.04 \\
 & \textbf{RANR} \citep{xu2024recurrent} & \cellcolor{green!30}94.15 & \cellcolor{green!30}82.90 & \cellcolor{green!30}83.53 & \cellcolor{green!30}91.66 & 42.69 & 74.15 & \cellcolor{orange!30}49.12 & 88.44 & 75.78 \\
& \textbf{G-ForgetNet} \citep{bohde2024markov} & \cellcolor{orange!30}78.09 & 63.84 & \cellcolor{orange!30}78.97 & \cellcolor{orange!30}91.79 & \cellcolor{green!30}86.70 & \cellcolor{green!30}88.80 & \cellcolor{green!30}54.73 & \cellcolor{green!30}95.09 & \cellcolor{green!30}82.89 \\
\midrule
\multirow{2}{*}{\textbf{GT}} 
& \textbf{RT} \citep{diao2022relational} & 50.01 & \cellcolor{orange!30}65.31 & 66.52 & 85.33 & \cellcolor{orange!30}83.20 & 65.33 & 32.52 & 84.55 & 66.18 \\
& \textbf{ET} \citep{muller2024towards}& 82.26 & 63.00 & 64.44 & 81.67 & 83.49 & 86.08 & \cellcolor{green!30}54.84 & 88.22 & \cellcolor{orange!30}80.13 \\
\midrule
\textbf{Ours} & \textbf{\fnet} & \cellcolor{cyan!30}\textbf{100.00} & \cellcolor{cyan!30}\textbf{91.60} & \cellcolor{cyan!30}\textbf{86.20} & \cellcolor{cyan!30}\textbf{93.17} & \cellcolor{cyan!30}\textbf{90.02} & \cellcolor{cyan!30}\textbf{98.60}& \cellcolor{cyan!30}\textbf{99.72} & \cellcolor{cyan!30}\textbf{99.68} & \cellcolor{cyan!30}\textbf{96.64}\\
\bottomrule
\end{tabular}
}
\end{table}

\subsection{Pairwise Relational Optimization: TSP}
TSP tests pair-state computation by mapping dense edge weights to a Hamiltonian-cycle edge set.

\paragraph{Settings.}
We use two benchmarks: a \emph{non-metric} TSP dataset of random complete graphs (up to 200 nodes, integer weights in $[1,100]$) and a \emph{metric} TSP dataset from random 2D integer coordinates. To handle solution multiplicity, we train within a DDPM framework \citep{ho2020denoising}. Reference optimal tours are obtained with Concorde, an exact branch-and-cut solver \citep{applegate1998solution}. The model is trained on $N \le 100$ and tested on $100 < N \le 200$ for OOD evaluation. We benchmark against Linkern, an implementation of the Chained Lin--Kernighan heuristic \citep{applegate2003chained}. We also evaluate a filtered subset to reduce ambiguity from multiple optimal tours. Full details are in Appendix~\ref{sec:appendix_tsp_details}.

\begin{figure*}[t]
    \centering
    \begin{subfigure}[b]{0.48\textwidth}
        \centering
        \includegraphics[width=\linewidth]{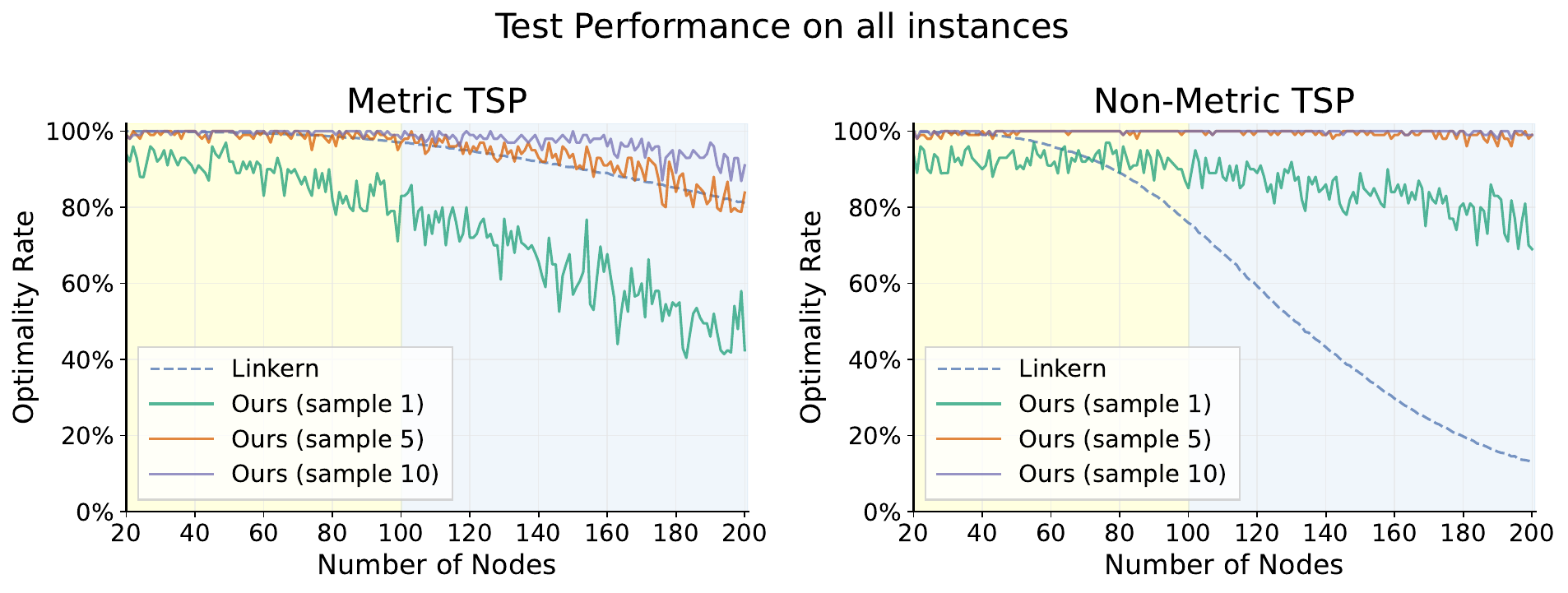}
        \caption{\textbf{Full Benchmark (Multi-Solution)}}
        \label{fig:tsp_a}
    \end{subfigure}
    \hfill
    \begin{subfigure}[b]{0.48\textwidth}
        \centering
        \includegraphics[width=\linewidth]{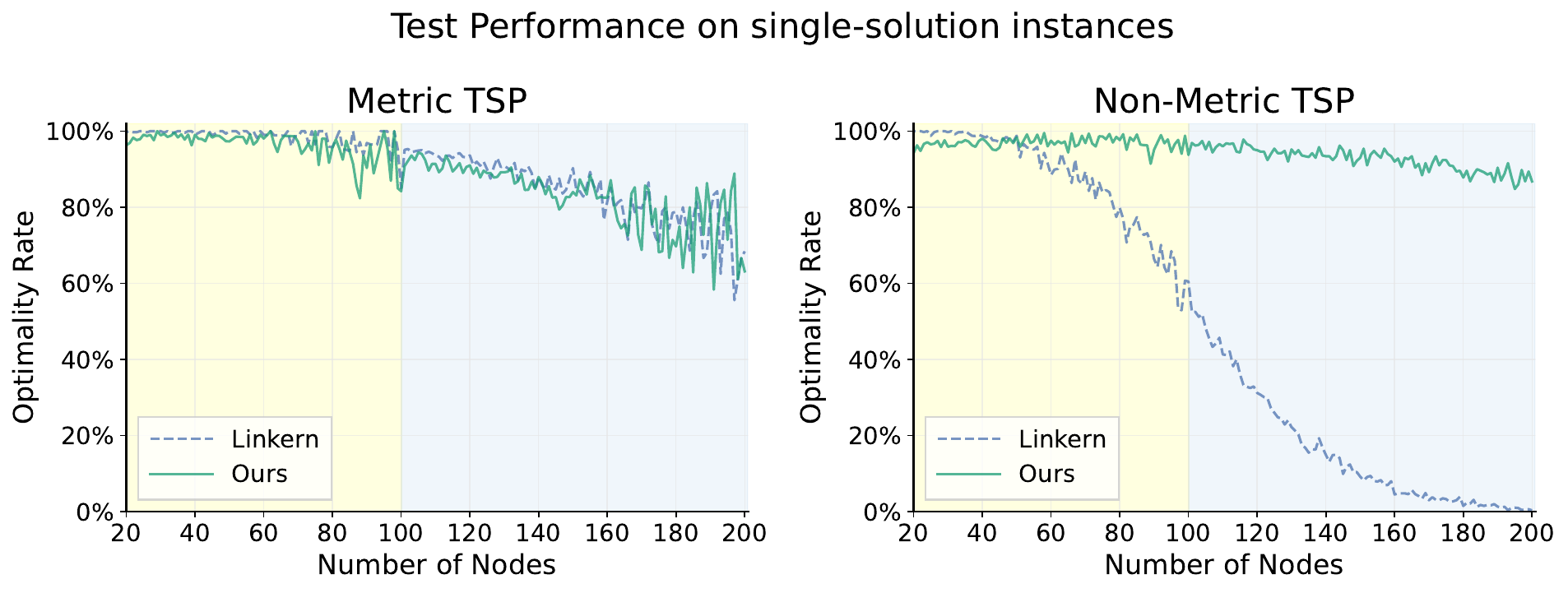}
        \caption{\textbf{Filtered Subset}}
        \label{fig:tsp_b}
    \end{subfigure}
    
    \vspace{1em} 

    \begin{subfigure}[b]{0.32\textwidth}
        \centering
        \includegraphics[width=\linewidth]{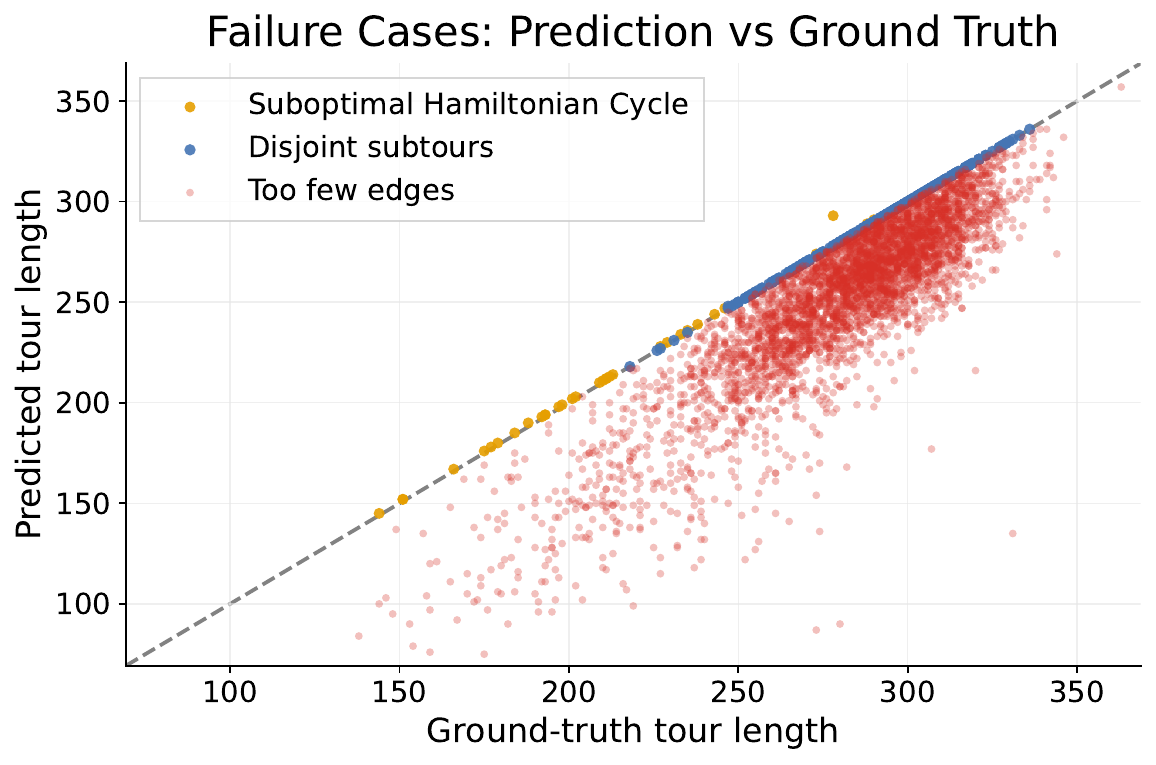}
        \caption{Failure Case Analysis}
        \label{fig:tsp_c}
    \end{subfigure}
    \hfill
    \begin{subfigure}[b]{0.32\textwidth}
        \centering
        \includegraphics[width=\linewidth]{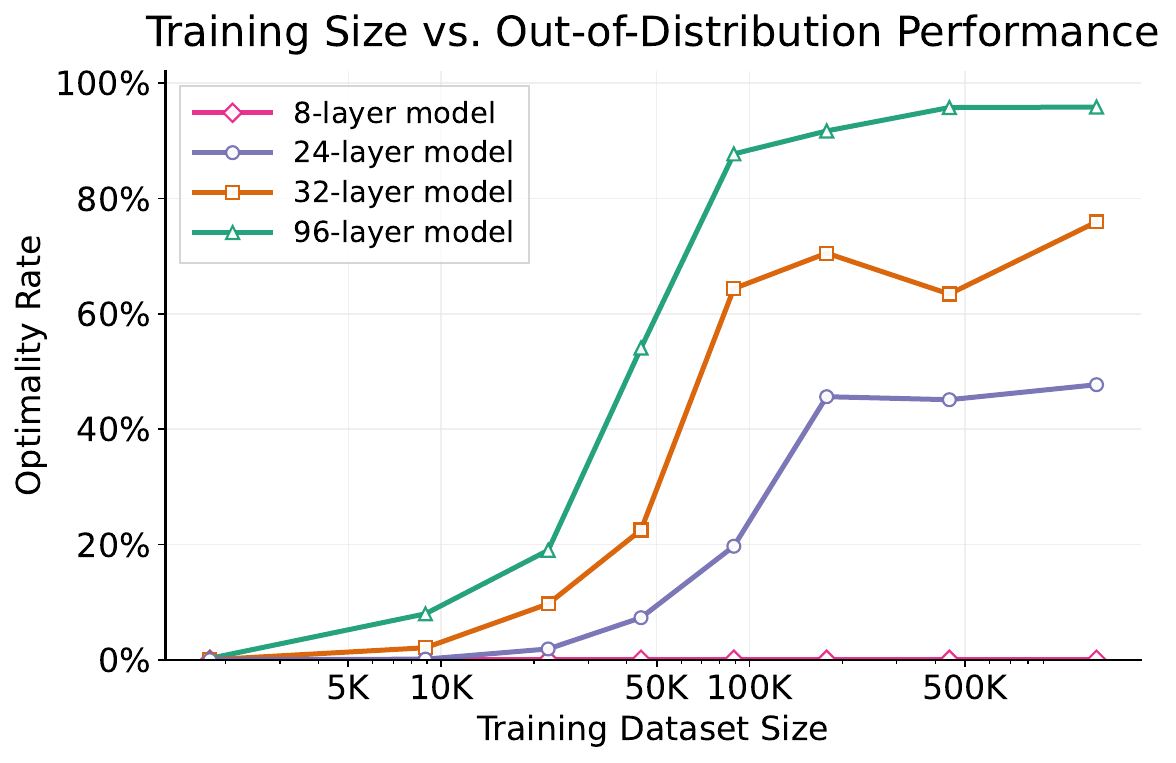}
        \caption{Data Scaling Analysis}
        \label{fig:tsp_d}
    \end{subfigure}
    \hfill
    \begin{subfigure}[b]{0.32\textwidth}
        \centering
        \includegraphics[width=\linewidth]{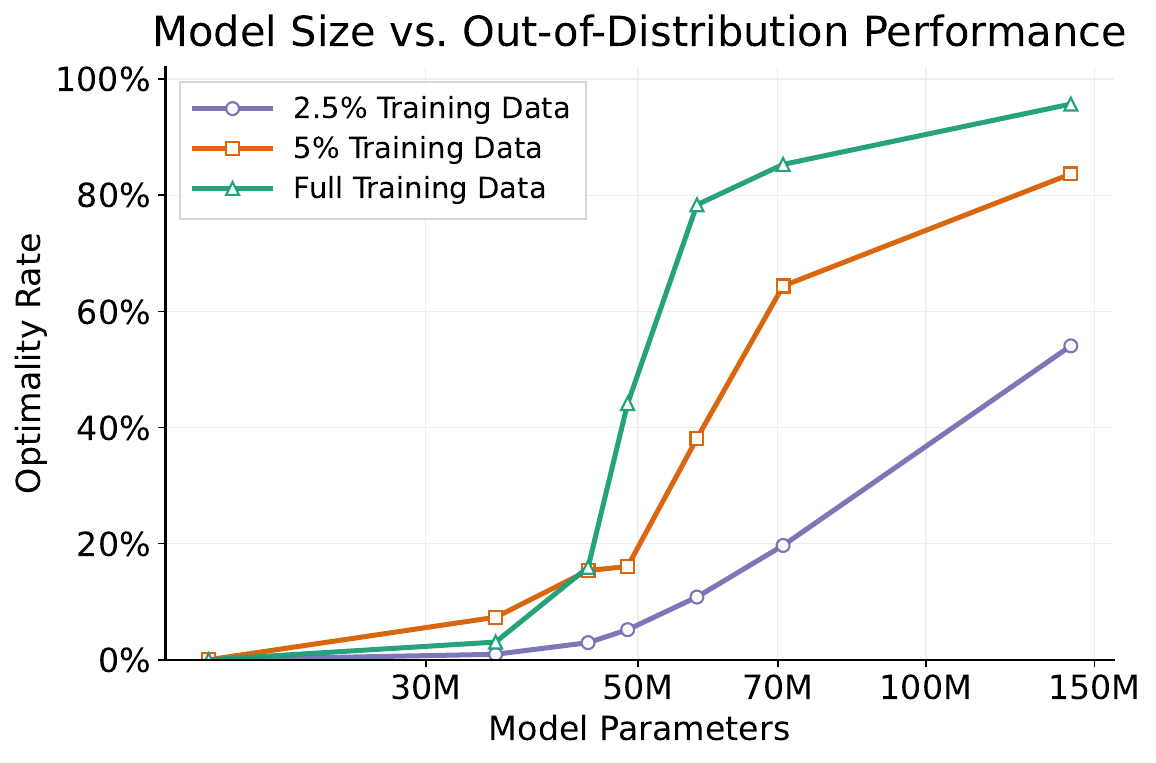}
        \caption{Model Scaling Analysis}
        \label{fig:tsp_e}
    \end{subfigure}
    \caption{\textbf{Analysis of \fnet on the TSP.} (a) Optimality rate on the benchmark (including multi-solution instances). The \textcolor{yellow!80!black}{\textbf{yellow region}} denotes validation on in-distribution sizes ($N \le 100$), while the \textcolor{cyan}{\textbf{cyan region}} demonstrates OOD generalization to larger problem sizes ($100 < N \le 200$). \textbf{(b)} Performance on the filtered subset, where perturb-and-resolve is used to reduce ambiguity from multiple optimal tours. \textbf{(c)} Distribution of error types. \textbf{(d-e)} Empirical trends with increasing training data and model depth under the evaluated configurations.}
    \label{fig:tsp_analysis}
\end{figure*}

\paragraph{Results.}
On held-out non-metric TSP instances ($100 < N \le 200$), \fnet achieves a $99.8\%$ optimality rate with 10 samples (\cref{fig:tsp_a}), versus Linkern's $38.8\%$. The gap widens with size: at $180 \le N \le 200$, Linkern drops to $16.1\%$ while \fnet maintains $99.4\%$. On the filtered non-metric subset (\cref{fig:tsp_b}), \fnet finds the exact optimum in $92.6\%$ of cases vs.\ Linkern's $15.7\%$; at $180 \le N \le 200$ this becomes $88.3\%$ vs.\ $1.3\%$. On metric TSP, where Linkern benefits from geometric structure, \fnet achieves comparable performance. Most failure cases produce incomplete tours rather than suboptimal ones, and this issue diminishes consistently with model scale (see Appendix~\ref{sec:appendix_tsp_details}).

\paragraph{Scaling Properties.}
As shown in \cref{fig:tsp_d,fig:tsp_e}, OOD performance improves with larger training sets and greater model depth (from $L=1$ to $L=96$) in the evaluated TSP configurations. In these configurations, deeper models perform better; every \fnet layer retains access to all candidate pivots.

\subsection{Transfer to Applied Relational Prediction}
Finally, we test whether the pair-relation inductive bias transfers beyond constructed algorithmic settings to molecular, contact-prediction, node-classification, and graph-level prediction benchmarks. The LRGB results in \Cref{tab:LRGB} show that, among the reported baselines, \fnet performs best on PCQM-Contact and COCO-SP, while remaining competitive on PascalVOC-SP and trailing Graph-Mamba on the two Peptides tasks. Results on ZINC are reported in Appendix \ref{sec:appendix_real_world}.

Training and validation curves for the remaining tasks are discussed in Appendix \ref{sec:appendix_real_world}. 

\begin{table}[h]
\centering
\caption{Mean performance on Long-Range Graph Benchmarks over five runs.}
\label{tab:LRGB}
\resizebox{\linewidth}{!}{
\setlength{\tabcolsep}{3pt}
\begin{tabular}{@{}llccccc}
\toprule
\multirow{3}{*}{\textbf{Class}} & \textbf{Task} & \textbf{Graph Class.} & \textbf{Graph Reg.} &  \textbf{Node Class.} &\textbf{Node Class.} &\textbf{Link Pred.} \\ 
& Dataset & Peptides-Func & Peptides-Struct &  PascalVOC-SP &COCO-SP &PCQM-Contact \\
& Metric & AP $\uparrow$ & MAE $\downarrow$ &  F1 $\uparrow$ &F1 $\uparrow$ &MRR $\uparrow$ \\ 
\midrule
\textbf{GNN} & GCN\citep{brossard2020graph} & 0.5930 & 0.3496 &  0.1268&0.0841 &0.3234 \\
\midrule
\multirow{2}{*}{\textbf{GT}} 
& Exphormer\citep{pmlr-v202-shirzad23a} & 0.6258 & 0.2512 &  0.3446&0.3430 &0.3587\\
& GPS\citep{rampavsek2022recipe} & 0.6535 & 0.2500 &  0.3724&0.3774&0.3437\\
\midrule
\textbf{SSM} & Graph-Mamba\citep{wang2024graph} & \cellcolor{cyan!20}0.6739 & \cellcolor{cyan!20}0.2478 &  \cellcolor{cyan!20}0.4191&0.3960&0.3395\\ 
\midrule
\textbf{Ours} & \textbf{\fnet} & \cellcolor{green!20}\textbf{0.6224}& \cellcolor{green!20}\textbf{0.2612}&  \cellcolor{green!20}\textbf{0.4046}&\cellcolor{cyan!20}\textbf{0.4901}&\cellcolor{cyan!20}\textbf{0.6143}\\
\bottomrule
\end{tabular}
}
\end{table}


\section{Related Work}

\subsection{Beyond 1-WL and Higher-Order GNNs}
Standard message-passing GNNs are bounded by the 1-WL test \citep{xu2019powerful,morris2019weisfeiler}. Approaches beyond this limit include higher-order invariant networks \citep{maron2019invariant,maron2019provably}, subgraph models \citep{bevilacqua2022equivariant,zhang2021nested}, and spectral methods \citep{zhang2024expressive,lim2023sign}. The k-Folklore Weisfeiler--Lehman hierarchy \citep{cai1992optimal,maron2019provably} is a useful yardstick for ordered-tuple models: it refines each tuple from the common-pivot multiset of its coordinate replacements, and FGNN explicitly mimics this refinement \citep{zhang2024beyond}. FloydNet is motivated independently by relation composition. Under atomic tuple initialization and invariant readout, the implemented \kfnet{k} is upper-bounded by k-FWL (see \cref{sec:appendix_proofs}); its pivot update is inspired by, but need not be injective like, the discrete k-FWL refinement.

\subsection{Entity, Pair, and Higher-Order Attention}
\textbf{Entity-state attention.} Graph Transformers apply Self-Attention to node states \citep{dwivedi2020generalization,ying2021transformers,kim2022pure,rampasek2022GPS,lai2023deeper}, commonly using positional or structural encodings to expose graph topology.

\textbf{Pair-state attention.} Explicit pair representations and triangle-indexed updates precede our work. Evoformer and Pairformer combine pair states with triangle multiplication and attention \citep{jumper2021highly,abramson2024accurate}; Edge Transformer applies triangular attention to node-pair states \citep{bergen2021edgetransformer}, with subsequent analysis relating it to the WL hierarchy \citep{muller2024towards}; EGT maintains an edge channel alongside node attention \citep{hussain2021egt}. These works use explicit pair states. PA differs in its attention candidate: for each pivot, both composable relations jointly construct the candidate key and value, which the target relation then scores.

\textbf{Higher-order attention.} \kfnet{k} extends PA to ordered tuple states through common-pivot coordinate replacement. Its attention-operation abstraction yields Self-Attention at $k=1$ and PA at $k=2$; the main differences from prior pair-state models are the jointly constructed pivot candidate and the ordered-tuple extension (\cref{sec:appendix_et_join}).

Appendix \cref{sec:appendix_extended_related} discusses NAR, combinatorial optimization, and message-passing limitations.

\section{Conclusion and Future Work}
\fnet updates explicit ordered relation states through pivot-specific relation composition. Its $k$-tuple formulation places this operation alongside entity-state attention and supports a k-FWL upper bound under the stated assumptions. Across BREC, CLRS, TSP, and graph prediction benchmarks, explicit relation states provide a useful inductive bias. The principal limitation is dense $\mathcal O(N^3)$ pair-pivot computation; future work will study sparse pivots, higher-order architectures, multimodal relational inputs, and connections to Abstract DP \citep{bertsekas2022abstract}.


\sloppy
\bibliography{references,references_extra}
\bibliographystyle{iclr2026_conference}

\newpage
\appendix
\renewcommand \thepart{} 
\renewcommand \partname{}
\part{Appendix} 
\parttoc 
\newpage

\section{Experimental Protocols and Dataset Details}
\label{sec:appendix_implementation} 

\subsection{General Implementation and Model Configurations} 
\label{sec:appendix_impl_and_conf}
Across all experiments, \fnet models were implemented in either PyTorch \citep{paszke2019pytorch} or JAX \citep{jax2018github}. The core architecture consists of a stack of $L$ \texttt{FloydBlock}s. Each block uses a multi-head Pivotal Attention layer followed by a standard MLP-based Feed-Forward Network (FFN). The FFN consists of two linear layers with a GELU activation in between. We use pre-layer normalization (LayerNorm) throughout the network.

We detail the architecture and training configurations for each benchmark dataset. Unless otherwise specified, we use a batch size of 1 for simplicity and set the AdamW momentum parameters to [0.9, 0.95].

\textbf{NED Counting.} For the NED counting dataset we use a model with $L=32$ layers, hidden representation dimension $d_r=128$, and 2 attention heads. Batch size = 64. Models are optimized with AdamW using an initial learning rate of $1\times10^{-4}$and a Reduce-on-Plateau learning-rate scheduler. The training objective is normalized mean absolute error (MAE) to match the baseline; early stopping is applied based on validation MAE. 

\textbf{BREC.} For the BREC dataset we use $L=32$ layers, $d_r=2$, and a single attention head. We replace layer norm with batch normalization, remove the feed-forward network (FFN) inside the pivotal-attention block, and perform all computations in \texttt{float64} precision. 

\textbf{CLRS.} For the CLRS benchmark we follow the official framework: \fnet is implemented as a processor while preserving the original encoder, decoder, and loss functions. For each algorithm, we select the deepest model that fits within our GPU memory constraints (layer counts vary with the algorithm’s hint size). Training uses AdamW with an initial learning rate of $1\times10^{-4}$, a short linear warmup, and cosine decay thereafter. Each algorithm is trained for up to $80\textrm{k}$ steps and evaluation is performed using the best validation checkpoint. Batch size, weight decay, and other hyperparameters follow the CLRS framework's default setting.

\textbf{TSP.} For the traveling salesman (TSP) experiments we use a larger model with $d_r=384$ and 6 attention heads. We train multiple models with layer depths ranging from $L=8$ to $L=96$ to study model scaling effects. The loss is binary cross-entropy applied to edge predictions. Optimization uses AdamW with a fixed learning rate $1\times 10^{-4}$ together with a Reduce-on-Plateau scheduler. Training is conducted in data-parallel mode across 64 GPUs, for 400 epochs, each consisting of 100 steps.

\textbf{Real-world tasks.} For real-world applications LRGB and ZINC, we use $L=48$, $d_{r}=384$, and 6 attention heads. Optimization follows the same recipe: AdamW with initial learning rate $1\times 10^{-4}$ and a Reduce-on-Plateau scheduler. 

\subsection{Dataset Details} 
\label{sec:appendix_data_detail}

\paragraph{NED Counting.}
This synthetic benchmark from \citet{zhang2024beyond} quantitatively probes GNN expressiveness through homomorphism-counting tasks. Models predict the frequencies of eight pattern graphs, including chordal-cycle patterns, at graph, node, and edge levels. The task construction uses Nested Ear Decomposition (NED) types to organize structural difficulty; performance therefore provides a benchmark-specific measure of counting capability rather than a universal measure of structural reasoning.

We observed that \fnet tends to drive the training loss to zero prematurely on the graph-level tasks , which in turn hampers generalization to the validation and test sets. To mitigate this, and following the original dataset generation protocol of \citet{chen2020can}, we regenerated a larger training set consisting of 50K randomly sampled graphs, which we used exclusively for the graph-level tasks. This adjustment stabilizes training and yields more reliable evaluation results, reducing the test error from 0.005–0.01 to nearly zero.

\paragraph{BREC.}
The Benchmark for Realized GNN Expressiveness (BREC) is designed to address shortcomings of earlier expressiveness benchmarks. It contains challenging graph pairs and measures the realized distinguishing power of an implemented model, enabling finer-grained comparison among models that all exceed the 1-WL baseline.

\begin{itemize}[noitemsep,topsep=0pt,leftmargin=*]
    \item The dataset contains 800 unique non-isomorphic graphs organized into 400 challenging pairs. These graphs are grouped into 4 distinct categories, each with a specific quantitative breakdown: 
    \begin{enumerate}[label=(\roman*),noitemsep]
        \item \textbf{Basic}: 60 pairs of simple, non-regular, 1-WL-indistinguishable graphs.
        \item \textbf{Regular}: 140 pairs, further subdivided into four types of regular graphs with increasing complexity (50 simple regular, 50 strongly regular, 20 4-vertex condition, and 20 distance regular pairs).
        \item \textbf{Extension}: 100 pairs designed to test specific theoretical GNN extensions (e.g., substructure-counting vs. k-hop subgraphs).
        \item \textbf{CFI}: 100 pairs generated using the Cai-Furer-Immerman construction, with their difficulty precisely calibrated against the WL hierarchy (specifically, 60 pairs distinguishable by 3-WL, 20 pairs by 4-WL, and 20 pairs that remain 4-WL-indistinguishable).
    \end{enumerate}
\end{itemize}

\paragraph{CLRS-30.}
The CLRS-30 benchmark \citep{velivckovic2022clrs} provides 30 selected algorithmic tasks based on the third edition of \textit{Introduction to Algorithms} \citep{cormen2009introduction}. In its size-generalization setting, processors train on smaller instances and are evaluated on larger ones. CLRS represents algorithm executions using input, output, and intermediate-hint probes attached to node, edge, or graph locations, enabling one processor architecture to be evaluated across diverse algorithms.

\section{Pivotal Attention, k-Floyd, and Architectural Details}

\label{sec:appendix_architecture}

\subsection{Pairwise Architecture, Readout, and Equivariance}
Let $G=(V,E,\mathbf X,\mathbf E,\mathbf G)$ be a finite attributed graph and treat $\mathbf E$ as a total ordered-pair input distinguishing edges, nonedges, and self-pairs when needed. The shared initialization and task decoders are
\begin{align}
\mR^{(0)}_{ij}&=\operatorname{Init}(\mathbf G,\mathbf X_i,\mathbf X_j,\mathbf E_{ij},\mathbf 1\{i=j\}),\nonumber\\
\widehat y_i&=D_{\mathrm{node}}(\mR^{(L)}_{is}),\quad
\widehat y_{ij}=D_{\mathrm{pair}}(\mR^{(L)}_{ij}),\quad
\widehat y_G=D_{\mathrm{graph}}(\mR^{(L)}_{ss}),
\label{eq:appendix_readouts}
\end{align}
where the optional SuperNode $s$ has a unique type and is fixed by permutations of original vertices. A FloydBlock applies the residual updates in \cref{eq:floydblock_attn,eq:floydblock_ffn}; normalization and FFN maps are shared over pair indices.

For a permutation $\pi$ of $V$, extend it by $\pi^+(s)=s$ and define
$(\pi^+\!\cdot\!\mR)_{ik}=\mR_{(\pi^+)^{-1}(i),(\pi^+)^{-1}(k)}$.
\begin{proposition}[Permutation equivariance, restated]
\label{prop:appendix_equivariance}
If inputs and masks are relabeled with the graph and all tuple maps are shared, then
$\operatorname{FloydBlock}(\pi^+\!\cdot\!\mR)=\pi^+\!\cdot\!\operatorname{FloydBlock}(\mR)$.
Consequently, node and pair predictions are equivariant and the graph prediction from $\mR_{ss}$ is invariant.
\end{proposition}
\begin{proof}
Relabeling reindexes the pivot sum through the bijection $p\mapsto\pi(p)$. Shared pointwise maps, softmax over pivots, residual connections, and normalization commute with this reindexing.
\end{proof}

\subsection{Multiplicative Pivotal Attention}
\label{sec:appendix_full_version}
The main experiments use addition to combine pair states associated with a pivot. Elementwise multiplication provides a bilinear alternative that can represent certain algebraic compositions. The following result concerns the value-composition submodule, not the subsequent softmax aggregation over multiple pivots.

\begin{proposition}[Matrix multiplication by an elementwise-product module]
\label{prop:matrix_product}
Let $\operatorname{vec}(A),\operatorname{vec}(B)\in\mathbb{R}^{9}$ be row-major column-vector encodings of $A,B\in\mathbb{R}^{3\times3}$. There exist linear maps $L_A,L_B:\mathbb{R}^{9}\to\mathbb{R}^{27}$ and $P:\mathbb{R}^{27}\to\mathbb{R}^{9}$ such that
\[
P\!\left(L_A\operatorname{vec}(A)\odot L_B\operatorname{vec}(B)\right)=\operatorname{vec}(AB).
\]
\end{proposition}
\begin{proof}
Index the 27 lifted coordinates by $(r,s,t)\in[3]^3$. Define
$[L_A\operatorname{vec}(A)]_{rst}=A_{rs}$ and
$[L_B\operatorname{vec}(B)]_{rst}=B_{st}$. Their elementwise product has coordinate $A_{rs}B_{st}$. Define $P$ by $[Pz]_{rt}=\sum_{s=1}^{3}z_{rst}$. Then
\[
[P(L_A\operatorname{vec}(A)\odot L_B\operatorname{vec}(B))]_{rt}
=\sum_{s=1}^{3}A_{rs}B_{st}=(AB)_{rt}.
\]
\end{proof}
For $A,B\in\mathrm{SO}(3)$, closure gives $AB\in\mathrm{SO}(3)$. If a pair state represents a transform between coordinate frames, the order of $A$ and $B$ is fixed by that frame convention. The proposition establishes exact representability of one pivot-conditioned product. Standard Pivotal Attention subsequently returns a normalized weighted sum over pivots; it equals a selected product only when one pivot is admissible or an explicit selector is imposed.

\subsection{Limitations and Engineering Optimization}
The primary trade-off of the \fnet architecture is its computational complexity. Pivotal Attention has $\mathcal{O}(N^3)$ cost in the present implementation, reflecting its dense all-pairs, pivot-enumerating computation. While this prohibits direct application to massive graphs (e.g., millions of nodes), it is well-suited for high-value reasoning tasks on dense graphs of moderate size (e.g., algorithmic reasoning, molecular modeling).

\begin{figure*}[h]
\centering
\includegraphics[width=\textwidth]{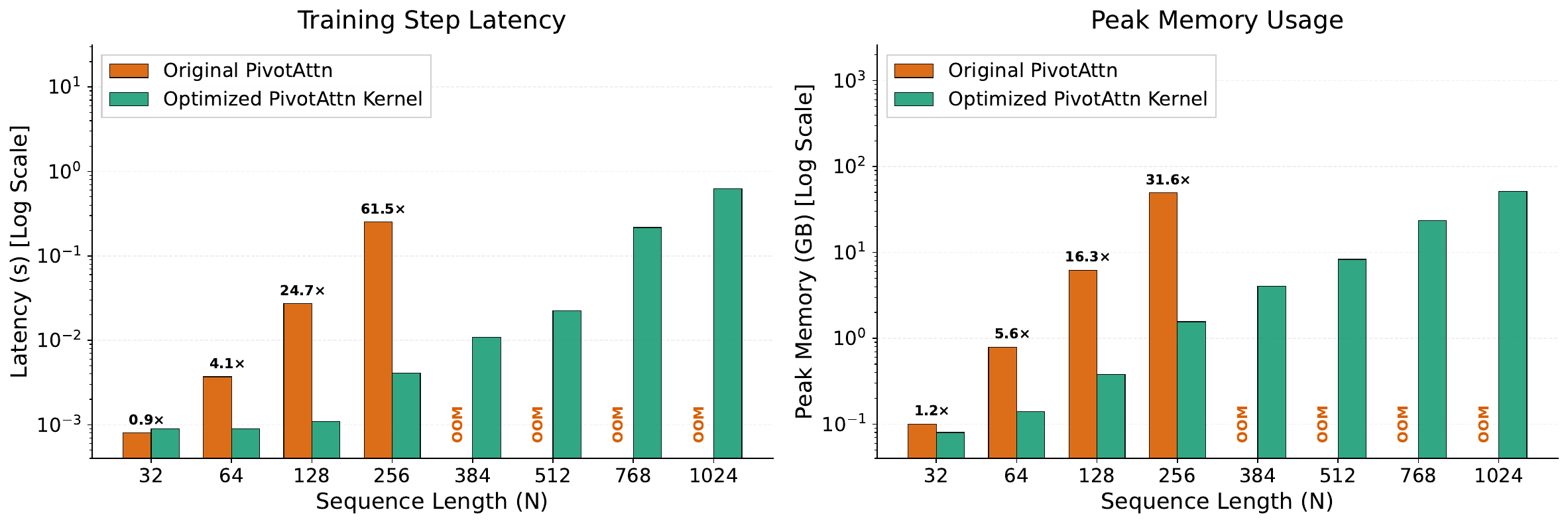}
\caption{\textbf{Performance Comparison of Original vs. Optimized PivotAttn Kernel.} Training step latency and peak GPU memory usage are shown as sequence length $N$ increases. The original implementation quickly leads to out-of-memory (OOM) errors. At $N=256$, the Original version requires 61.5x the runtime and 31.6x the peak memory of the Optimized version.} \label{fig:kernel_speedup}
\end{figure*}

To mitigate this, we implemented a highly optimized CUDA kernel and an optimized PyTorch version at \url{https://anonymous.4open.science/r/FloydNet}, which reduce memory usage from $\mathcal{O}(N^3)$ to $\mathcal{O}(N^2)$ and improve wall-clock speed by over $20\times$. This allows efficient training on graphs up to $N=256$ on standard GPUs. Extending \fnet to larger, sparse graphs via techniques such as sparse attention is a promising direction for future work. \cref{fig:kernel_speedup} provides a detailed comparison of the runtime and GPU memory consumption between the kernel-optimized and the original implementations as the sequence length increases. 

\subsection{Higher-Order Relation States in k-Floyd}
\label{sec:appendix_floyd_plus} 
\begin{figure}[t]
\begin{minipage}{0.44\textwidth}
    \includegraphics[width=1.0\textwidth]{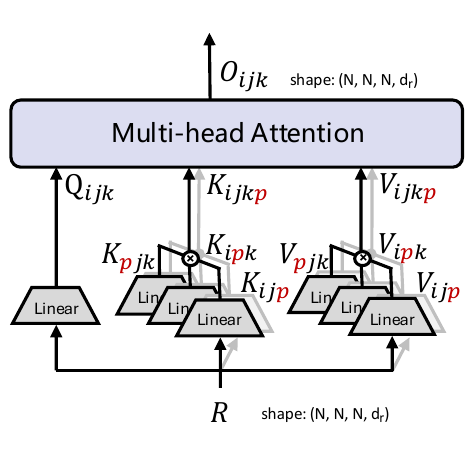}
\end{minipage}
\hfill
\begin{minipage}{0.56\textwidth}
\centering
\begin{lstlisting}[language=python, mathescape, breaklines=true]
def PivotAttnCore(q_ijk, k_pjk, k_ipk, k_ijp, 
                            v_pjk, v_ipk, v_ijp):
    # All input tensors of shape [b,n,n,n,h,d_h]
    # come from different linear transform of R

    # Additive role-specific composition
    k_ijkp = (k_pjk[:, None, :, :, :, :, :] +
              k_ipk[:, :, None, :, :, :, :] +
              k_ijp[:, :, :, None, :, :, :])
    v_ijkp = (v_pjk[:, None, :, :, :, :, :] +
              v_ipk[:, :, None, :, :, :, :] +
              v_ijp[:, :, :, None, :, :, :])
    q_ijkp = q_ijk[:, :, :, :, None, :, :]

    a_ijkp = torch.sum(q_ijkp * k_ijkp, dim=-1)
    d = q_ijk.shape[-1] # per head embed dim
    a_ijkp = a_ijkp / (d ** 0.5) # [b,n,n,n,n,h]
    w_ijkp = torch.softmax(a_ijkp, dim=-2) # dim p
    
    o_ijkp = v_ijkp * w_ijkp[..., None]
    o_ijk = torch.sum(o_ijkp, dim=-3) # dim p
    return o_ijk # [b,n,n,n,h,d_h], same as input
\end{lstlisting}
\end{minipage}
\caption{\textbf{Pivotal Attention for \kfnet{3}.} The state of an ordered triple $(i,j,k)$ attends over pivots $p$ using the three coordinate-replacement states.}
\label{fig:pirvotal_attention_3_floyd}
\end{figure}

The \fnet architecture extends to \kfnet{k}, whose states are indexed by ordered $k$-tuples in $V^k$. Increasing $k$ enlarges the tuple state space and the family of coordinate-replacement interactions available to the model; the formal k-FWL upper bound is given in \cref{thm:fwl_upper_bound}.

\paragraph{Initialization of Higher-Order Relationship Tensors.}
A \kfnet{k} model uses an initial tuple-state tensor $\mR^{(0)}\in\mathbb{R}^{N^k\times d_r}$, indexed by $\mathbf v\in V^k$. In general, initialization is a shared map of the available ordered lower-order features within $\mathbf v$, including equality indicators and edge/nonedge types. For a simple attributed graph, one possible choice is
\begin{equation}
\mR^{(0)}_{\mathbf v}=\operatorname{Init}_k\!\left(
\mathbf G,
(\mathbf X_{v_a})_{a\in[k]},
(\mathbf E_{v_a v_b})_{a,b\in[k]},
(\mathbf{1}\{v_a=v_b\})_{a,b\in[k]}
\right).
\label{eq:k_tuple_init}
\end{equation}
This ordered construction covers directed pair features, repeated coordinates, and nonedges. Task-specific higher-order attributes may be concatenated when available.
\paragraph{Pivotal Attention.}
The figure shows the additive $k=3$ specialization of \cref{eq:k_pivotal_messages}. Relative to the pairwise case, tuple dimensionality increases and each pivot message combines one role-specific projection for every coordinate replacement:
\begin{enumerate}[topsep=0pt,leftmargin=*]
    \item \textbf{Tensor Dimensionality:} The \textbf{Query}, \textbf{Key}, and \textbf{Value} tensors are promoted from operating on pairs to operating on k-tuples. For instance, in a \kfnet{3} model, these tensors would have a shape of $(b, n, n, n, h, d_h)$, where $b$ is the batch size and $n$ is the number of nodes.
    \item \textbf{Aggregation Logic:} For a target tuple $\mathbf v=(v_1,\ldots,v_k)$ and pivot $p\in V$, the message is formed from the $k$ coordinate-replacement states $\mR_{\mathbf v[r\leftarrow p]}$, $r\in[k]$, and attention is normalized over $p$.
\end{enumerate}

\paragraph{Implementations.}
For the BREC benchmark, we developed a proof-of-concept implementation of \kfnet{k} to evaluate the higher-order construction empirically. Developing optimized kernels for these variants remains future work.

\subsection{Complexity of k-Floyd Pivotal Attention}
\label{sec:appendix_complexity}
For fixed $k$ and $h$, dense tuple states, and an FFN width of $\Theta(d_r)$, the projections and FFN cost $\mathcal O(N^k d_r^2)$, while pivot scoring, normalization, and value aggregation cost $\mathcal O(N^{k+1}d_r)$. Hence a block takes
\begin{equation}
\mathcal O(N^{k+1}d_r+N^k d_r^2)
\end{equation}
time. A naive implementation materializes $\mathcal O(N^{k+1}d_r)$ pivot-conditioned activations. For $k=2$, the fused kernel avoids this materialization and uses $\mathcal O(N^2d_r)$ forward intermediate storage, excluding implementation-dependent workspace and saved training activations.

\subsection{Relation-State Comparison with Transformer-Style Models}

\subsubsection{Relationship to Edge Transformer: Join-V vs. Join-K}
\label{sec:appendix_et_join}
Prior pair-state architectures use several triangle-indexed update factorizations \citep{jumper2021highly,abramson2024accurate,bergen2021edgetransformer}. PA specifically constructs each pivot candidate key and value jointly from $\mR_{ij}$ and $\mR_{jk}$ before the target $\mR_{ik}$ scores it. This construction aligns with the common-pivot organization used in our k-FWL upper-bound analysis.

To clarify this broader state-space perspective, \Cref{tab:arch_comparison} compares standard entity-state attention, pair-state \fnet, and higher-order \kfnet{k}. The current work establishes an operation-level abstraction: common-pivot replacement yields Self-Attention at $k=1$ and relational attention at higher orders. It does not claim identity between complete architectures; rather, it provides a basis for future architectures that integrate sequence and relational computation more fully.

General graph canonization is polynomial-time equivalent to graph isomorphism; despite the quasipolynomial-time GI algorithm of \citet{babai2016graph}, no polynomial-time algorithm is known for either problem \citep{grohe2017descriptive}. Graph Transformers commonly use positional or structural encodings to expose topology without imposing a canonical node order \citep{ying2021transformers,rampavsek2022recipe}.

\fnet takes a complementary approach by operating on an all-pairs relationship tensor. Permuting node indices induces the corresponding permutation of both tensor axes, and shared projections plus symmetric pivot aggregation preserve this equivariance. The architecture therefore represents relational structure without requiring a canonical node order, at the cost of lifting computation to a higher-dimensional state space.

\begin{table}[h!]
\centering
\caption{Architectural properties. Expressivity entries for FloydNet are upper bounds under the assumptions of \cref{thm:fwl_upper_bound}.}
\label{tab:arch_comparison}
\small
\setlength{\tabcolsep}{4pt}
\begin{tabular}{@{}cccc@{}}
\toprule
\textbf{Property} & \textbf{Transformer} & \textbf{\fnet} & \textbf{k-\fnet} \\
\midrule
\makecell[ct]{Core Mechanism} & \makecell[t]{Self-Attention} & \makecell[t]{Pivotal Attention} & \makecell[t]{Higher-order \\ Pivotal Attention} \\
\rule{0pt}{3ex}
\makecell[lt]{Operating Space} & \makecell[t]{Node States \\ ($\mathcal{O}(N d)$)} & \makecell[t]{Ordered Pair States \\ ($\mathcal{O}(N^2 d)$)} & \makecell[t]{Ordered k-Tuple States \\ ($\mathcal{O}(N^k d)$)} \\
\rule{0pt}{3ex}
\makecell[lt]{Interaction Cost} & \makecell[t]{$\mathcal{O}(N^2 d)$} & \makecell[t]{$\mathcal{O}(N^3 d)$} & \makecell[t]{$\mathcal{O}(N^{k+1}d)$} \\
\rule{0pt}{3ex}
\makecell[lt]{Permutation Equivariant} & \makecell[t]{Without index PEs} & \makecell[t]{Yes} & \makecell[t]{Yes} \\
\rule{0pt}{3ex}
\makecell[lt]{Structural Encoding} & \makecell[t]{Architecture- \\ dependent} & \makecell[t]{Optional} & \makecell[t]{Optional} \\
\rule{0pt}{3ex}
\makecell[lt]{Expressivity Bound} & \makecell[t]{Encoding- \\ dependent} & \makecell[t]{$\preceq$ 2-FWL} & \makecell[t]{$\preceq$ k-FWL} \\
\bottomrule
\end{tabular}
\end{table}

\subsection{Structured-Compute Scaling as a Complementary Axis}
\label{sec:appendix_scaling}

Our TSP experiments show a consistent empirical trend: OOD performance improves as training data and model depth increase (\cref{fig:tsp_d,fig:tsp_e}). This observation is narrower than a general scaling law, but it motivates a complementary view of model scaling. Whereas large sequence models often increase capacity primarily through parameter count and data \citep{kaplan2020scaling}, FloydNet also increases the structured relational computation performed for each instance through depth and dense pair-pivot interaction.

We hypothesize that the effective capacity of relational architectures depends not only on parameter count but also on the order, depth, and amount of structured computation applied to explicit relation states. The k-Floyd abstraction makes this a natural future research axis: entity-, pair-, and higher-order attention can be studied within one state-space framework while varying computational order and depth. Establishing whether this route benefits broader deliberative tasks requires experiments beyond the present TSP scaling study.

\section{Extended Discussion on Related Work}
\label{sec:appendix_extended_related}

\subsection{Neural Algorithmic Reasoning}
\label{sec:appendix_related_nar}
The broader effort to learn or execute algorithms with neural networks predates the modern NAR terminology, including learning-to-execute models, differentiable memory, and Neural GPUs \citep{zaremba2014learning,graves2014neural,kaiser2015neural}. Graph-specific neural execution later brought this agenda into the relational domain \citep{velivckovic2019neural}. The NAR framing \citep{velivckovic2021neural} and the CLRS benchmark \citep{velivckovic2022clrs} consolidated this line of work around algorithm trajectories, graph representations, and size generalization. \citet{dudzik2022graph} relate GNN computation to dynamic programming, while \citet{ibarz2022generalist} introduce Triplet-GMPNN and demonstrate multi-task learning across all 30 CLRS algorithms with one processor. Subsequent work examines out-of-distribution generalization \citep{mahdavi2022towards} and scalability to larger instances \citep{minder2023salsa}.

Many CLRS-oriented systems, including several baselines considered here, use message-passing processors and intermediate hint supervision. These are influential design choices rather than defining requirements of NAR. \fnet instead uses global all-pairs refinement; under our reported protocol, its no-hint variants achieve strong performance on many algorithms, showing that Pivotal Attention can support algorithmic learning without intermediate supervision in these settings.

\subsection{Machine Learning for Combinatorial Optimization}
\label{sec:appendix_related_co}
Applying neural networks to combinatorial optimization (CO), particularly the Traveling Salesman Problem (TSP), has attracted growing attention \citep{bengio2021machine}. Autoregressive methods construct solutions step by step using attention-based decoders \citep{kool2018attention,kwon2020pomo}. Some non-autoregressive approaches, including diffusion-based methods, instead predict and iteratively refine solution structure \citep{sun2023difusco}, achieving strong results on large-scale metric TSP through denoised edge probability maps. However, most existing neural TSP solvers focus on the metric (Euclidean) case and rely on problem-specific architectural designs or coordinate-based features that exploit geometric structure.

In contrast, our approach is designed and evaluated on the more challenging \emph{general} (non-metric) TSP, where edge weights are arbitrary and no geometric heuristics apply. By learning directly on the dense distance matrix through pairwise relation-state refinement, \fnet tests whether the same relation-composition mechanism can operate without coordinate-based geometric features.

\subsection{Over-Smoothing and Over-Squashing}
\label{sec:appendix_related_oversquashing}
The local nature of message passing introduces two well-known pathologies. Over-smoothing causes node representations to converge as depth increases \citep{li2018deeper}, while over-squashing compresses exponentially growing information into fixed-width feature vectors when long-range communication is needed \citep{Alon2021On, topping2022understanding}. \citet{digiovanni2023oversquashing} formally characterize how over-squashing limits the function classes that MPNNs can learn.

\fnet avoids the specific sequential, finite-hop communication pattern that underlies these analyses of local MPNNs: each pair state has direct access to compositions through every candidate pivot. As formalized in \cref{prop:dependency_depth}, an individual value-composition tree can double its number of base pair-state leaves with each layer. This global access is designed to mitigate distance-driven information bottlenecks, but it does not imply immunity to representation homogenization, finite-width information loss, or optimization difficulties.

\section{k-FWL Upper Bound and Its Scope}
\label{sec:appendix_proofs}

This section formalizes how the common-pivot k-Floyd construction sits within the folklore WL hierarchy. The result is an invariance statement: k-FWL-equivalent tuples remain indistinguishable to shared k-Floyd updates.

\begin{theorem}[k-FWL upper bound, restated]
\label{thm:appendix_fwl_upper_bound}
Assume attributed atomic tuple initialization, tuple maps shared over indices, permutation-invariant pivot aggregation, and an invariant graph readout using only final tuple states. If two ordered tuples have the same k-FWL color after $\ell$ rounds, then their \kfnet{k} states after $\ell$ layers are equal. Consequently, \kfnet{k} cannot distinguish graph pairs that k-FWL fails to distinguish. In particular, \fnet is upper-bounded by 2-FWL, equivalently ordinary 3-WL under the folklore convention below.
\end{theorem}

Let $G=(V,E)$ be an attributed graph and let $\mathbf v=(v_1,\ldots,v_k)\in V^k$ be an ordered tuple. Its \emph{attributed atomic type}, denoted $\operatorname{at}_G(\mathbf v)$, records the graph-level attribute $\mathbf G$ (when supplied to initialization), the equality pattern of its coordinates, their ordered node attributes, and the edge/nonedge attributes of every ordered coordinate pair. For $r\in[k]$ and $p\in V$, write $\mathbf v[r\leftarrow p]$ for coordinate replacement as in \cref{sec:general_fnet}.

\begin{definition}[Folklore k-WL]
\label{def:kfwl}
The initial k-FWL color is $c_G^{(0)}(\mathbf v)=\operatorname{at}_G(\mathbf v)$. At round $\ell$, define the pivot multiset
\begin{equation}
M_G^{(\ell)}(\mathbf v)=
\left\{\!\!\left\{
\big(c_G^{(\ell)}(\mathbf v[1\leftarrow p]),\ldots,
 c_G^{(\ell)}(\mathbf v[k\leftarrow p])\big):p\in V
\right\}\!\!\right\},
\label{eq:kfwl_multiset}
\end{equation}
and update
\begin{equation}
c_G^{(\ell+1)}(\mathbf v)=
\operatorname{HASH}\!\left(c_G^{(\ell)}(\mathbf v),M_G^{(\ell)}(\mathbf v)\right),
\label{eq:kfwl_update}
\end{equation}
where $\operatorname{HASH}$ is injective in its two arguments. Two graphs are distinguished after $L$ rounds if the multisets of colors over $V^k$ differ. This folklore convention has the distinguishing power of ordinary $(k+1)$-WL \citep{cai1992optimal,maron2019provably}.
\end{definition}

\begin{lemma}[Layer preservation]
\label{lem:layer_preservation}
Under the assumptions of \cref{thm:appendix_fwl_upper_bound}, if
$c_G^{(\ell)}(\mathbf v)=c_H^{(\ell)}(\mathbf w)$, then
$\mR_{G,\mathbf v}^{(\ell)}=\mR_{H,\mathbf w}^{(\ell)}$.
\end{lemma}
\begin{proof}
We use induction on $\ell$. At $\ell=0$, equal atomic types produce equal states because initialization is a shared function of atomic type. Suppose the claim holds at round $\ell$. Equality at round $\ell+1$ implies, by injectivity of $\operatorname{HASH}$, both
$c_G^{(\ell)}(\mathbf v)=c_H^{(\ell)}(\mathbf w)$ and equality of the pivot multisets in \cref{eq:kfwl_multiset}. The induction hypothesis maps each replacement-color vector in these equal multisets to an equal ordered vector of neural states. Therefore the corresponding pivot-conditioned keys, values, and target queries are equal up to a bijective reordering of pivots. Softmax and its weighted sum are invariant to that reordering, so Pivotal Attention produces equal outputs. Shared output projections, residual connections, normalization, and pointwise FFNs preserve equality, proving the claim.
\end{proof}

\begin{proof}[Proof of \cref{thm:appendix_fwl_upper_bound}]
Apply \cref{lem:layer_preservation} through all layers. If k-FWL does not distinguish $G$ and $H$, their final tuple-color multisets agree; the lemma implies that the corresponding neural-state multisets agree. Any invariant readout satisfying assumption (iv) therefore returns the same graph representation on $G$ and $H$.
\end{proof}

\begin{remark}[Why the upper bound need not be generally tight]
\label{rem:softmax_noninjective}
The argument above uses only permutation invariance and does not require attention to be injective. Indeed, normalized attention is not generally multiplicity-sensitive. If every pivot supplies the same key $\boldsymbol\kappa$ and value $\boldsymbol\nu$, then for any query and any number $m$ of such pivots,
\[
\sum_{p=1}^{m}
\frac{\exp(\langle\mathbf q,\boldsymbol\kappa\rangle)}
{\sum_{r=1}^{m}\exp(\langle\mathbf q,\boldsymbol\kappa\rangle)}
\boldsymbol\nu=\boldsymbol\nu.
\]
Thus duplicating identical pivot items may change the k-FWL multiset while leaving the softmax output unchanged. Exact k-FWL simulation would require additional assumptions or a position-aware, multiplicity-sensitive injective aggregator. We do not attribute that property to the evaluated softmax architecture.
\end{remark}

\begin{proposition}[Value-composition depth, restated]
\label{prop:appendix_dependency_depth}
Fix one target pair and pivot choice per layer and expand only the two value-side states composed at each chosen pivot. After $L$ FloydBlocks, this binary tree has at most $2^L$ initial pair-state leaves.
\end{proposition}
\begin{proof}[Proof of \cref{prop:appendix_dependency_depth}]
The selected value-composition tree has one leaf at depth zero. At each subsequent layer, its chosen pivot value composes two trees from the previous layer. If each contains at most $2^{L-1}$ leaves, their parent contains at most $2^L$. This argument deliberately excludes query/key dependencies and aggregation over other pivots, which belong to the full attention dependency graph rather than to one binary value-composition tree.
\end{proof}

\section{Role-Specific Experimental Details}

\subsection{BREC: Pair-Level Concordance}
\label{sec:appendix_full_result_brec} 
To thoroughly evaluate the expressive power of our framework, we provide a detailed analysis on BREC. The results, presented in \Cref{tab:brec_full}, compare our models against theoretical tests and other representative GNN architectures.
Under its RPC protocol, our base model, \fnet, distinguishes exactly the same graph pairs as the 3-WL reference: all 60 Basic and 100 Extension pairs, together with the same 50/140 Regular and 60/100 CFI pairs. This exact pair-level concordance is stronger than equality of aggregate scores and matches the 3-WL reference success set on BREC. It remains a statement about realized expressiveness on this finite benchmark, not a proof of equivalence on arbitrary graphs.

This agreement identifies a complementary pattern on BREC rather than establishing that \fnet inherits every limitation of 3-WL. Subgraph-based models such as KP-GNN perform strongly on the evaluated regular-graph subset, while \fnet performs better on the reported CFI subset. The result motivates testing whether their respective features can be combined.

To verify this, we conducted a hybrid experiment. By augmenting \fnet with pre-computed graph features from the KP-GNN framework (denoted \fnet-KP), the model's performance on regular graphs improves significantly. As shown in \Cref{tab:brec_full}, the hybrid model correctly distinguishes 106 regular pairs, which matches KP-GNN's strength in this area while retaining \fnet's power on CFI graphs. This boosts the total accuracy from 67.5\% to 81.5\%, demonstrating that the pairwise relation-state model can be combined effectively with local structural features.

For \kfnet{k}, \kfnet{3} resolves all regular graph pairs and 80 CFI pairs, while \kfnet{4} achieves 99.8\% accuracy by distinguishing 99 of the 100 CFI pairs. At each tuple order, the success set agrees pair by pair with the corresponding WL reference on BREC; we do not infer the behavior of higher-dimensional WL tests from the remaining pair.
\begin{table*}[h!]
\centering
\caption{Detailed pair-distinguishing results on BREC. Under this protocol, the \kfnet{k} scores agree with the displayed WL references. Results are also shown for representative GNN architectures.}
\label{tab:brec_full}
\resizebox{\textwidth}{!}{%
\begin{tabular}{@{}llccccc@{}}
\toprule
\multirow{2}{*}{\textbf{Type}} & \multirow{2}{*}{\textbf{Model}} & \textbf{Basic} & \textbf{Regular} & \textbf{Extension} & \textbf{CFI} & \textbf{Total} \\
& & (acc. on 60) & (acc. on 140) & (acc. on 100) & (acc. on 100) & (acc. on 400) \\ \midrule
\multirow{2}{*}{Theoretical Tests} & 1-WL Test & 0 & 0 & 0 & 0 & 0 (0.0\%) \\
& 3-WL Test & \textbf{60} & 50 & \textbf{100} & 60 & 270 (67.5\%) \\ \midrule
MPNNs& GIN (MPNN) & 0 & 0 & 0 & 0 & 0 (0.0\%) \\ \midrule
\color{black}  \multirow{3}{*}{Graph Transformers}& Graphormer & 16 & 12 & 41 & 10 & 79 (19.8\%) \\
 & \color{black}GDT+RWSE& \color{black}57& \color{black}50& \color{black}96& \color{black}0&\color{black}203 (50.8\%)\\
 & \color{black}PPGT& \color{black}\textbf{60}& \color{black}50& \color{black}\textbf{100}& \color{black}24&\color{black}234 (58.5\%)\\ \midrule
\multirow{2}{*}{Subgraph GNNs} & I$^2$-GNN & \textbf{60} & 100 & \textbf{100} & 21 & 281 (70.2\%) \\
& KP-GNN & \textbf{60} & \textbf{106} & 98 & 11 & 275 (68.8\%) \\ \midrule
\multirow{2}{*}{k-WL GNNs} & PPGN & 60 & 50 & 100 & 23 & 233 (58.2\%) \\
& KC-SetGNN & 60 & 50 & 100 & 1 & 211 (52.8\%) \\ \midrule
\multirow{2}{*}{\textbf{Ours}} & \textbf{\fnet} & \cellcolor{green!20}\textbf{60} & \cellcolor{green!20}50 & \cellcolor{green!20}\textbf{100} & \cellcolor{green!20}60 & \cellcolor{green!20}270 (67.5\%) \\
& \textbf{\fnet-KP}\textsuperscript{*} & \cellcolor{cyan!20}\textbf{60} & \cellcolor{cyan!20}\textbf{106} & \cellcolor{cyan!20}\textbf{100} & \cellcolor{cyan!20}\textbf{60} & \cellcolor{cyan!20}\textbf{326 (81.5\%)} \\
& \textbf{\kfnet{3}} & \cellcolor{cyan!20}\textbf{60} & \cellcolor{cyan!20}\textbf{140} & \cellcolor{cyan!20}\textbf{100} & \cellcolor{cyan!20}\textbf{80} & \cellcolor{cyan!20}\textbf{380 (95.0\%)} \\
& \textbf{\kfnet{4}} & \cellcolor{cyan!20}\textbf{60} & \cellcolor{cyan!20}\textbf{140} & \cellcolor{cyan!20}\textbf{100} & \cellcolor{cyan!20}\textbf{99} & \cellcolor{cyan!20}\textbf{399 (99.8\%)} \\
\bottomrule
\multicolumn{7}{l}{\textsuperscript{*}\footnotesize{\fnet using KP-GNN's pre-computed graph features.}}
\end{tabular}%
}
\end{table*}

\subsection{CLRS: Algorithmic Relation Composition}
\label{sec:appendix_full_result_clrs} 
\paragraph{Hint vs. No Hint.}
In the CLRS benchmark, \emph{hints} are intermediate algorithm states used as an additional supervision signal. In the standard CLRS baseline setup, hint-supervised processors recurrently predict these states across algorithmic steps \citep{velivckovic2022clrs}. In our experiments, hints improve maximum subarray, LCS length, matrix-chain order, optimal BST, Activity Selector, and DFS, but do not improve most other tasks (\cref{tab:clrs_hint_vs_nohint}). Hint prediction also increases memory use, so the hint-augmented models use $L=8$, whereas no-hint models use $L=32$ to $L=80$; comparisons therefore reflect the evaluated configurations rather than an intrinsic disadvantage of hints. For no-hint training, we remove intermediate targets and use one processor step per FloydBlock. We additionally use a SALSA-inspired extrapolation setting: models train on $N=16$ and are evaluated at $N=160$, one of the larger scales considered by SALSA-CLRS \citep{minder2023salsa}. As reported in \cref{tab:clrs_vs_salsa}, the no-hint configuration attains high accuracy on BFS, Dijkstra, and MST at this size, while the evaluated hint-augmented configuration performs less well.
\paragraph{Online vs. Offline.}
The CLRS framework provides both a preprocessed, finite training dataset and an on-the-fly data sampling procedure. We denote training with the finite dataset as \emph{offline training} and training with unlimited sampled data as \emph{online training}. As discussed earlier, \fnet exhibits strong scaling capability, but with limited training data it tends to suffer from premature convergence, which in turn harms generalization. We trained both offline and online variants using identical hyperparameters, denoting the online version as \fnet$_{full}$ in \cref{tab:test_results_epic}. Across nearly all algorithms, online training yields superior OOD test performance. For instance, in sorting tasks, performance improves from nearly perfect to consistently $100\%$, while in KMP matching, accuracy rises dramatically from $27\%$ to $99\%$.
\begin{table*}[h]
\centering
\caption{Early-stopped validation accuracy (\%) on all 30 CLRS algorithms. }
\label{tab:val_results_all}
\resizebox{\linewidth}{!}{%
\begin{tabular}{lccccccc} \toprule
\textbf{Algorithm} & \textbf{Deep Sets} & \textbf{GAT} & \textbf{Memnet} & \textbf{MPNN} & \textbf{PGN} & \textbf{\fnet} & \textbf{\fnet$_{full}$}\\
\midrule
Activity Selector & $ 83.50\% \pm  0.17$ & $ 92.40\% \pm  0.50$ & $ 34.59\% \pm  2.15$ & $ 93.89\% \pm  0.39$ & $ 82.26\% \pm  0.19$ & \cellcolor{green!30}$97.20\%\pm 0.87$ & \cellcolor{cyan!30}$100.00\%\pm 0.00$\\
Articulation Points & $ 99.63\% \pm  0.31$ & \cellcolor{cyan!30}$ 100.00\% \pm  0.00$ & $ 16.84\% \pm  1.03$ & \cellcolor{cyan!30}$ 100.00\% \pm  0.00$ & \cellcolor{cyan!30}$ 100.00\% \pm  0.00$ & \cellcolor{cyan!30}$100.00\%\pm 0.00$ & \cellcolor{cyan!30}$100.00\%\pm 0.00$\\
Bellman-Ford & $ 81.12\% \pm  0.14$ & $ 99.28\% \pm  0.14$ & $ 68.75\% \pm  0.42$ & \cellcolor{green!30}$ 99.48\% \pm  0.05$ & $ 99.35\% \pm  0.05$ & $99.47\%\pm 0.61$ & \cellcolor{cyan!30}$100.00\%\pm 0.00$\\
BFS & \cellcolor{cyan!30}$ 100.00\% \pm  0.00$ & \cellcolor{cyan!30}$ 100.00\% \pm  0.00$ & $ 70.70\% \pm  0.09$ & \cellcolor{cyan!30}$ 100.00\% \pm  0.00$ & \cellcolor{cyan!30}$ 100.00\% \pm  0.00$ & \cellcolor{cyan!30}$100.00\%\pm 0.00$ & \cellcolor{cyan!30}$100.00\%\pm 0.00$\\
Binary Search & $ 93.34\% \pm  0.41$ & \cellcolor{green!30}$ 95.72\% \pm  0.17$ & $ 20.33\% \pm  0.28$ & $ 94.19\% \pm  0.12$ & $ 94.17\% \pm  0.08$ & $94.67\%\pm 0.74$ & \cellcolor{cyan!30}$99.47\%\pm 0.67$\\
Bridges & $ 99.36\% \pm  0.05$ & \cellcolor{cyan!30}$ 100.00\% \pm  0.00$ & $ 96.46\% \pm  1.13$ & \cellcolor{cyan!30}$ 100.00\% \pm  0.00$ & \cellcolor{cyan!30}$ 100.00\% \pm  0.00$ & \cellcolor{cyan!30}$100.00\%\pm 0.00$ & \cellcolor{cyan!30}$100.00\%\pm 0.00$\\
Bubble Sort & $ 81.51\% \pm  1.02$ & \cellcolor{green!30}$ 95.44\% \pm  1.01$ & $ 92.64\% \pm  0.14$ & $ 94.53\% \pm  1.84$ & $ 87.17\% \pm  5.46$ & \cellcolor{cyan!30}$100.00\%\pm 0.00$ & \cellcolor{cyan!30}$100.00\%\pm 0.00$\\
DAG Shortest Paths & $ 92.25\% \pm  0.28$ & $ 96.81\% \pm  0.05$ & $ 81.90\% \pm  0.05$ & \cellcolor{green!30}$ 99.93\% \pm  0.05$ & $ 99.80\% \pm  0.00$ & $99.80\%\pm 0.00$ & \cellcolor{cyan!30}$100.00\%\pm 0.00$\\
DFS & $ 62.76\% \pm  1.26$ & $ 99.22\% \pm  0.64$ & $ 47.72\% \pm  0.45$ & \cellcolor{cyan!30}$ 100.00\% \pm  0.00$ & \cellcolor{cyan!30}$ 100.00\% \pm  0.00$ & \cellcolor{cyan!30}$100.00\%\pm 0.00$ & \cellcolor{cyan!30}$100.00\%\pm 0.00$\\
Dijkstra & $ 80.34\% \pm  0.42$ & $ 99.22\% \pm  0.40$ & $ 67.38\% \pm  0.70$ & \cellcolor{green!30}$ 99.67\% \pm  0.14$ & $ 99.28\% \pm  0.05$ & \cellcolor{cyan!30}$100.00\%\pm 0.00$ & \cellcolor{cyan!30}$100.00\%\pm 0.00$\\
Find Max. Subarray & $ 91.41\% \pm  0.22$ & $ 95.00\% \pm  0.32$ & $ 27.91\% \pm  0.08$ & $ 95.13\% \pm  0.37$ & $ 95.30\% \pm  0.16$ & \cellcolor{green!30}$95.63\%\pm 1.42$ & \cellcolor{cyan!30}$99.90\%\pm 0.00$\\
Floyd-Warshall & $ 35.79\% \pm  0.04$ & $ 87.28\% \pm  0.09$ & $ 31.29\% \pm  0.04$ & $ 89.14\% \pm  0.03$ & $ 88.70\% \pm  0.15$ & \cellcolor{green!30}$98.97\%\pm 0.06$ & \cellcolor{cyan!30}$99.87\%\pm 0.06$\\
Graham Scan & $ 87.66\% \pm  0.24$ & $ 97.85\% \pm  0.11$ & $ 53.53\% \pm  1.58$ & $ 98.45\% \pm  0.15$ & $ 89.06\% \pm  0.27$ & \cellcolor{green!30}$99.20\%\pm 0.28$ & \cellcolor{cyan!30}$100.00\%\pm 0.00$\\
Heapsort & $ 81.84\% \pm  0.33$ & $ 87.24\% \pm  2.23$ & $ 54.04\% \pm  0.28$ & $ 94.27\% \pm  0.11$ & $ 90.36\% \pm  0.67$ & \cellcolor{green!30}$98.10\%\pm 0.00$ & \cellcolor{cyan!30}$100.00\%\pm 0.00$\\
Insertion Sort & $ 89.58\% \pm  0.28$ & $ 95.18\% \pm  0.58$ & $ 94.40\% \pm  0.14$ & $ 96.74\% \pm  0.19$ & $ 84.57\% \pm  0.82$ & \cellcolor{green!30}$99.50\%\pm 0.42$ & \cellcolor{cyan!30}$100.00\%\pm 0.00$\\
Jarvis' March & $ 72.82\% \pm  0.42$ & \cellcolor{green!30}$ 98.38\% \pm  0.16$ & $ 37.92\% \pm  6.61$ & $ 97.94\% \pm  0.25$ & $ 88.34\% \pm  0.36$ & $98.10\%\pm 0.00$ & \cellcolor{cyan!30}$100.00\%\pm 0.00$\\
KMP Matcher & $ 98.03\% \pm  0.21$ & $ 99.76\% \pm  0.08$ & $ 9.67\% \pm  0.00$ & \cellcolor{green!30}$ 99.87\% \pm  0.05$ & $ 94.14\% \pm  0.99$ & \cellcolor{cyan!30}$100.00\%\pm 0.00$ & \cellcolor{cyan!30}$100.00\%\pm 0.00$\\
LCS Length & $ 69.24\% \pm  0.36$ & $ 77.00\% \pm  0.19$ & $ 67.69\% \pm  0.24$ & $ 77.88\% \pm  0.42$ & $ 69.19\% \pm  0.04$ & \cellcolor{cyan!30}$100.00\%\pm 0.00$ & \cellcolor{cyan!30}$100.00\%\pm 0.00$\\
Matrix Chain Order & $ 94.46\% \pm  0.02$ & $ 99.37\% \pm  0.03$ & $ 93.91\% \pm  0.10$ & $ 99.12\% \pm  0.04$ & $ 99.21\% \pm  0.03$ & \cellcolor{green!30}$99.50\%\pm 0.10$ & \cellcolor{cyan!30}$99.93\%\pm 0.06$\\
Minimum & $ 97.59\% \pm  0.11$ & \cellcolor{green!30}$ 97.74\% \pm  0.21$ & $ 95.56\% \pm  0.10$ & $ 97.64\% \pm  0.05$ & $ 97.07\% \pm  0.14$ & $99.50\%\pm 0.10$ & \cellcolor{cyan!30}$100.00\%\pm 0.00$\\
MST-Kruskal & $ 83.79\% \pm  2.01$ & $ 97.93\% \pm  0.25$ & $ 64.65\% \pm  0.95$ & \cellcolor{green!30}$ 99.71\% \pm  0.17$ & $ 99.12\% \pm  0.08$ & \cellcolor{cyan!30}$100.00\%\pm 0.00$ & \cellcolor{cyan!30}$100.00\%\pm 0.00$\\
MST-Prim & $ 74.61\% \pm  0.32$ & $ 98.37\% \pm  0.14$ & $ 74.09\% \pm  0.28$ & \cellcolor{green!30}$ 99.02\% \pm  0.09$ & $ 97.79\% \pm  0.14$ & $98.30\%\pm 2.94$ & \cellcolor{cyan!30}$100.00\%\pm 0.00$\\
Na\"{i}ve String Match & \cellcolor{cyan!30}$ 100.00\% \pm  0.00$ & \cellcolor{cyan!30}$ 100.00\% \pm  0.00$ & $ 9.91\% \pm  0.20$ & \cellcolor{cyan!30}$ 100.00\% \pm  0.00$ & $ 50.33\% \pm  0.08$ & \cellcolor{cyan!30}$100.00\%\pm 0.00$ & \cellcolor{cyan!30}$100.00\%\pm 0.00$\\
Optimal BST & $ 92.02\% \pm  0.14$ & $ 93.30\% \pm  0.49$ & $ 90.86\% \pm  0.40$ & $ 93.88\% \pm  0.11$ & $ 93.20\% \pm  0.27$ & \cellcolor{green!30}$96.30\%\pm 1.27$ & \cellcolor{cyan!30}$98.53\%\pm 1.23$\\
Quickselect & $ 42.30\% \pm  0.92$ & $ 83.82\% \pm  1.86$ & $ 6.56\% \pm  0.25$ & $ 88.74\% \pm  0.78$ & $ 54.02\% \pm  0.17$ & \cellcolor{green!30}$99.60\%\pm 0.00$ & \cellcolor{cyan!30}$99.97\%\pm 0.06$\\
Quicksort & $ 79.69\% \pm  1.12$ & $ 92.97\% \pm  0.40$ & $ 93.16\% \pm  0.24$ & $ 95.70\% \pm  0.40$ & $ 54.30\% \pm  1.42$ & \cellcolor{cyan!30}$100.00\%\pm 0.00$ & \cellcolor{cyan!30}$100.00\%\pm 0.00$\\
Segments Intersect & $ 77.49\% \pm  0.12$ & $ 90.82\% \pm  0.16$ & $ 71.57\% \pm  1.08$ & $ 93.84\% \pm  0.20$ & $ 78.32\% \pm  0.18$ & \cellcolor{green!30}$96.07\%\pm 0.23$ & \cellcolor{cyan!30}$99.87\%\pm 0.12$\\
SCC & $ 89.52\% \pm  1.23$ & \cellcolor{cyan!30}$ 100.00\% \pm  0.00$ & $ 70.57\% \pm  1.43$ & \cellcolor{cyan!30}$ 100.00\% \pm  0.00$ & $ 99.93\% \pm  0.05$ & \cellcolor{cyan!30}$100.00\%\pm 0.00$ & \cellcolor{cyan!30}$100.00\%\pm 0.00$\\
Task Scheduling & $ 99.16\% \pm  0.04$ & $ 99.80\% \pm  0.04$ & $ 84.80\% \pm  0.09$ & \cellcolor{cyan!30}$ 100.00\% \pm  0.00$ & $ 99.06\% \pm  0.08$ & \cellcolor{cyan!30}$100.00\%\pm 0.00$ & \cellcolor{cyan!30}$100.00\%\pm 0.00$\\
Topological Sort & $ 47.23\% \pm  0.81$ & \cellcolor{cyan!30}$ 100.00\% \pm  0.00$ & $ 8.30\% \pm  0.50$ & \cellcolor{cyan!30}$ 100.00\% \pm  0.00$ & \cellcolor{cyan!30}$ 100.00\% \pm  0.00$ & \cellcolor{cyan!30}$100.00\%\pm 0.00$ & \cellcolor{cyan!30}$100.00\%\pm 0.00$\\
\midrule
Overall average & $ 80.93\%$ & $ 95.66\%$ & $ 57.92\%$ & $ 96.63\%$ & $ 89.47\%$ & \cellcolor{green!30}$99.00\%$ & \cellcolor{cyan!30}$99.92\%$\\
\bottomrule
\end{tabular}
}
\end{table*}
\begin{table*}[t]
\centering
\caption{Test performance (\%) of all models on all 30 CLRS algorithms, reported as mean $\pm$ std over 3 runs.}
\label{tab:test_results_epic}
\resizebox{\linewidth}{!}{%
\begin{tabular}{lcccccc}
\toprule
\textbf{Algorithm} & \textbf{Triplet-GMPNN} & \textbf{RT} & \textbf{G-ForgetNet} & \textbf{RANR} & \textbf{\fnet} & \textbf{\fnet$_{full}$} \\
\midrule
Activity Selector      & $95.18\%\pm 0.45$ & $87.72\%\pm 2.7$ & \cellcolor{cyan!30}\textbf{$99.03\%\pm 0.10$} & $95.23\%\pm 0.71$ & $92.27\%\pm 0.80$ & \cellcolor{green!30}$95.30\%\pm 1.56$ \\
Articulation Points    & $91.04\%\pm 0.92$ & $34.15\%\pm 14.6$ & \cellcolor{green!30}$97.97\%\pm 0.46$ & $26.32\%\pm{27.34}$ & $93.03\%\pm 1.37$ & \cellcolor{cyan!30}$100.00\%\pm 0.00$ \\
Bellman-Ford           & $97.39\%\pm 0.19$ & $94.24\%\pm 1.5$ & \cellcolor{green!30}$99.18\%\pm 0.11$ & $96.00\%\pm{0.38}$ & $94.10\%\pm 0.79$ & \cellcolor{cyan!30}$100.00\%\pm 0.00$ \\
BFS                    & $99.93\%\pm 0.03$ & $99.14\%\pm 0.7$ & \cellcolor{green!30}$99.96\%\pm 0.01$ & \cellcolor{cyan!30}$100.0\%\pm{0.0}$ & \cellcolor{cyan!30}$100.00\%\pm 0.00$ & \cellcolor{cyan!30}$100.00\%\pm 0.00$ \\
Binary Search          & $77.58\%\pm 2.35$ & $81.48\%\pm 6.7$ & \cellcolor{green!30}$85.96\%\pm 1.59$ & $64.71\%\pm{6.79}$ & $75.53\%\pm 5.40$ & \cellcolor{cyan!30}$94.47\%\pm 5.26$ \\
Bridges                & $97.70\%\pm 0.34$ & $37.88\%\pm 11.8$ & \cellcolor{green!30}$99.43\%\pm 0.15$ & $72.22\%\pm{12.66}$ & $99.45\%\pm 0.21$ & \cellcolor{cyan!30}$100.00\%\pm 0.00$ \\
Bubble Sort            & $80.51\%\pm 9.10$ & $38.22\%\pm 13.0$ & $83.19\%\pm 2.59$ & \cellcolor{green!30}$95.78\%\pm{0.40}$ & $89.70\%\pm 17.41$ & \cellcolor{cyan!30}$100.00\%\pm 0.00$ \\
DAG Shortest Paths     & $98.19\%\pm 0.30$ & $96.61\%\pm 1.6$ & \cellcolor{green!30}$99.37\%\pm 0.03$ & $96.40\%\pm{1.47}$ & $97.90\%\pm 0.00$ & \cellcolor{cyan!30}$100.00\%\pm 0.00$ \\
DFS                    & \cellcolor{cyan!30}$100.0\%\pm 0.00$ & $39.23\%\pm 10.5$ & $74.31\%\pm 5.03$ & \cellcolor{cyan!30}$100.0\%\pm{0.0}$ & \cellcolor{cyan!30}$100.00\%\pm 0.00$ & \cellcolor{cyan!30}$100.00\%\pm 0.00$ \\
Dijkstra               & $96.05\%\pm 0.60$ & $91.20\%\pm 5.8$ & \cellcolor{cyan!30}$99.14\%\pm 0.06$ & $95.04\%\pm{1.62}$ & $97.07\%\pm 2.22$ & \cellcolor{green!30}$97.67\%\pm 2.55$ \\
Find Max. Subarray     & $76.36\%\pm 0.43$ & $66.52\%\pm 3.7$ & $78.97\%\pm 0.70$ & \cellcolor{green!30}$83.53\%\pm{2.17}$ & $68.60\%\pm 2.69$ & \cellcolor{cyan!30}$86.20\%\pm 0.95$ \\
Floyd-Warshall         & $48.52\%\pm 1.04$ & $31.59\%\pm 7.6$ & $56.32\%\pm 0.86$ & $27.49\%\pm{6.95}$ & \cellcolor{green!30}$81.40\%\pm 2.21$ & \cellcolor{cyan!30}$96.20\%\pm 1.93$ \\
Graham Scan            & $93.62\%\pm 0.91$ & $74.15\%\pm 7.4$ & \cellcolor{green!30}$97.67\%\pm 0.14$ & $76.20\%\pm{4.51}$ & $96.20\%\pm 0.00$ & \cellcolor{cyan!30}$99.63\%\pm 0.15$\\
Heapsort               & $49.13\%\pm 10.35$ & $32.96\%\pm 14.8$ & $57.47\%\pm 6.08$ & $93.07\%\pm{1.03}$ & \cellcolor{green!30}$99.00\%\pm 0.00$ & \cellcolor{cyan!30}$100.00\%\pm 0.00$ \\
Insertion Sort         & $87.21\%\pm 2.80$ & $89.43\%\pm 9.0$ & \cellcolor{green!30}$98.40\%\pm 0.21$ & $93.00\%\pm{1.77}$ & $95.35\%\pm 0.35$ & \cellcolor{cyan!30}$100.00\%\pm 0.00$ \\
Jarvis' March          & $91.01\%\pm 1.30$ & \cellcolor{green!30}$94.57\%\pm 2.2$ & $88.53\%\pm 2.96$ & $91.83\%\pm{1.77}$ & \cellcolor{cyan!30}$99.00\%\pm 0.00$ & \cellcolor{cyan!30}$99.50\%\pm 0.14$ \\
KMP Matcher            & $19.51\%\pm 4.57$ & $0.03\%\pm 0.1$ & $12.45\%\pm 3.12$ & $4.54\%\pm{2.60}$ & \cellcolor{green!30}$27.37\%\pm 9.08$ & \cellcolor{cyan!30}$99.43\%\pm 0.38$ \\
LCS Length             & $80.51\%\pm 1.84$ & $83.32\%\pm 4.1$ & \cellcolor{green!30}$85.43\%\pm 0.47$ & $66.91\%\pm{2.53}$ & $91.70\%\pm 4.61$ & \cellcolor{cyan!30}$95.70\%\pm 1.60$ \\
Matrix Chain Order     & $91.68\%\pm 0.59$ & \cellcolor{cyan!30}$91.89\%\pm 1.2$ & $91.08\%\pm 0.51$ & $25.12\%\pm{1.86}$ & $89.23\%\pm 0.64$ & $91.50\%\pm 1.41$\\
Minimum                & $98.43\%\pm 0.01$ & $95.28\%\pm 2.0$ & \cellcolor{green!30}$99.26\%\pm 0.08$ & $96.92\%\pm{0.09}$ & $99.07\%\pm 0.06$ & \cellcolor{cyan!30}$100.00\%\pm 0.00$ \\
MST-Kruskal            & $89.93\%\pm 0.43$ & $64.91\%\pm 11.8$ & $91.25\%\pm 0.40$ & $67.29\%\pm{0.93}$ & \cellcolor{green!30}$93.47\%\pm 1.36$ & \cellcolor{cyan!30}$99.63\%\pm 0.21$ \\
MST-Prim               & $87.64\%\pm 1.79$ & $85.77\%\pm 7.9$ & \cellcolor{green!30}$95.19\%\pm 0.33$ & $86.60\%\pm{4.42}$ & $90.53\%\pm 12.05$ & \cellcolor{cyan!30}$99.67\%\pm 0.31$ \\
Na\"{i}ve String Match & $78.67\%\pm 4.99$ & $65.01\%\pm 32.3$ & \cellcolor{green!30}$97.02\%\pm 0.77$ & $93.71\%\pm{2.26}$ & \cellcolor{cyan!30}$100.00\%\pm 0.00$ & \cellcolor{cyan!30}$100.00\%\pm 0.00$ \\
Optimal BST            & $73.77\%\pm 1.48$ & $74.40\%\pm 2.6$ & \cellcolor{cyan!30}$83.58\%\pm 0.49$ & $36.04\%\pm{12.55}$ & $78.00\%\pm 1.13$ & \cellcolor{green!30}$82.87\%\pm 0.96$ \\
Quickselect            & $0.47\%\pm 0.25$ & $19.18\%\pm 17.3$ & $6.30\%\pm 0.85$ & \cellcolor{cyan!30}$87.08\%\pm{2.21}$ & \cellcolor{green!30}$83.70\%\pm 6.36$ & $80.33\%\pm 8.05$ \\
Quicksort              & $85.69\%\pm 4.53$ & $39.42\%\pm 13.2$ & $73.28\%\pm 6.25$ & $94.73\%\pm{0.63}$ & \cellcolor{green!30}$99.80\%\pm 0.10$ & \cellcolor{cyan!30}$100.00\%\pm 0.00$ \\
Segments Intersect     & $97.64\%\pm 0.09$ & $84.94\%\pm 2.6$ & \cellcolor{green!30}$99.06\%\pm 0.39$ & $97.30\%\pm{0.29}$ & $95.87\%\pm 0.12$ & \cellcolor{cyan!30}$99.90\%\pm 0.10$ \\
SCC                    & $43.43\%\pm 3.15$ & $28.59\%\pm 15.2$ & $53.53\%\pm 2.48$ & $48.43\%\pm{8.01}$ & \cellcolor{green!30}$89.27\%\pm 1.75$ & \cellcolor{cyan!30}$90.07\%\pm 0.57$ \\
Task Scheduling        & $87.25\%\pm 0.35$ & $82.93\%\pm 1.8$ & $84.55\%\pm 0.35$ & $88.08\%\pm{1.30}$& \cellcolor{green!30}$87.23\%\pm 7.40$ & \cellcolor{cyan!30}$91.05\%\pm 1.75$\\
Topological Sort       & $87.27\%\pm 2.67$ & $80.62\%\pm 17.5$ & \cellcolor{green!30}$99.92\%\pm 0.02$ & $74.00\%\pm{8.18}$ & \cellcolor{cyan!30}$100.00\%\pm 0.00$ & \cellcolor{cyan!30}$100.00\%\pm 0.00$ \\
\midrule
\textbf{Overall average} & $80.04\%$ & $66.18\%$ & $82.89\%$ & $75.78\%$ & \cellcolor{green!30}$90.13\%$ & \cellcolor{cyan!30}$96.64\%$\\
\bottomrule
\end{tabular}
}
\end{table*}
\begin{table*}[t!]
\centering
\small
\caption{Comparison of \fnet$_{full}$ performance with and without hints on the CLRS-30 test set. Algorithms are grouped by the impact of hints. The \textbf{$\Delta$} column shows the accuracy point difference (With Hint - No Hint), with positive (beneficial) changes in green and negative (detrimental) changes in red. OOM indicates an Out-of-Memory error.}
\label{tab:clrs_hint_vs_nohint}
\sisetup{table-format=-2.2, tight-spacing} 
\begin{tabular}{l S S S}
\toprule
\textbf{Algorithm} & {\textbf{With Hint (\%)}} & {\textbf{No Hint (\%)}} & {\textbf{$\Delta$}} \\
\midrule
\multicolumn{4}{l}{\textit{\textbf{Group 1: Algorithms Where Hints Improve Performance}}} \\
\midrule
Find Max. Subarray   & 86.20 & 80.57 & \cellcolor{green!25} +5.63 \\
LCS Length           & 95.70 & 86.10 & \cellcolor{green!25} +9.60 \\
Matrix Chain Order   & 91.50 & 85.15 & \cellcolor{green!25} +6.35 \\
Optimal BST          & 82.87 & 74.50 & \cellcolor{green!25} +8.37 \\
Activity Selector    & 95.30 & 93.40 & \cellcolor{green!25} +1.90 \\
DFS                  & 100.00 & 75.10 & \cellcolor{green!25} +24.90 \\
\addlinespace
\midrule
\multicolumn{4}{l}{\textit{\textbf{Group 2: Algorithms Where Hints Degrade Performance or Cause OOM}}} \\
\midrule
Binary Search        & 24.70 & 94.47 & \cellcolor{red!25} -69.77 \\
Floyd-Warshall       & 64.20 & 96.20 & \cellcolor{red!25} -32.00 \\
Graham Scan          & 88.80 & 99.63 & \cellcolor{red!25} -10.83 \\
Quickselect          & 52.60 & 80.33 & \cellcolor{red!25} -27.73 \\
Segments Intersect   & 79.10 & 99.90 & \cellcolor{red!25} -20.80 \\
Bellman-Ford         & 99.60 & 100.00 & \cellcolor{red!25} -0.40 \\
DAG Shortest Paths   & 96.80 & 100.00 & \cellcolor{red!25} -3.20 \\
Dijkstra             & 97.67 & 99.40 & \cellcolor{red!25} -1.73 \\
\cdashline{1-4}[1pt/1pt]
Articulation Points  & {OOM} & 100.00 & \cellcolor{red!25} {---} \\
Bridges              & {OOM} & 100.00 & \cellcolor{red!25} {---} \\
Bubble Sort          & {OOM} & 100.00 & \cellcolor{red!25} {---} \\
Heapsort             & {OOM} & 100.00 & \cellcolor{red!25} {---} \\
Jarvis' March        & {OOM} & 99.50 & \cellcolor{red!25} {---} \\
KMP Matcher          & {OOM} & 99.43 & \cellcolor{red!25} {---} \\
MST-Kruskal          & {OOM} & 99.63 & \cellcolor{red!25} {---} \\
Quicksort            & {OOM} & 100.00 & \cellcolor{red!25} {---} \\
SCC                  & {OOM} & 90.07 & \cellcolor{red!25} {---} \\
\addlinespace
\midrule
\multicolumn{4}{l}{\textit{\textbf{Group 3: Algorithms with Saturated or Neutral Performance}}} \\
\midrule
BFS                  & 100.00 & 100.00 & 0.00 \\
Insertion Sort       & 99.50 & 100.00 & -0.50 \\
Minimum              & 99.90 & 100.00 & -0.10 \\
MST-Prim             & 99.67 & 99.65 & +0.02 \\
Na\"{i}ve String Match & 100.00 & 100.00 & 0.00 \\
Task Scheduling      & 91.05 & 91.50 & -0.45 \\
Topological Sort     & 100.00 & 99.75 & +0.25 \\
\bottomrule
\end{tabular}
\end{table*}

\begin{table}[htbp]
\centering
\caption{Node accuracy on Erdős-Rényi (ER) random graphs ($N=160$). We compare standard inference vs. hint-aided inference (H). \textbf{Bold} indicates the best performance.}
\label{tab:clrs_vs_salsa}
\small
\renewcommand{\arraystretch}{1.2} 
\setlength{\tabcolsep}{5pt} 
\begin{tabular}{lcccccccc}
\toprule
\multirow{2}{*}{\textbf{Model}} & \multicolumn{2}{c}{\textbf{BFS}} & \multicolumn{2}{c}{\textbf{DFS}} & \multicolumn{2}{c}{\textbf{Dijkstra}} & \multicolumn{2}{c}{\textbf{MST}} \\
\cmidrule(lr){2-3} \cmidrule(lr){4-5} \cmidrule(lr){6-7} \cmidrule(lr){8-9}
 & Std. & Hint & Std. & Hint & Std. & Hint & Std. & Hint \\ 
\midrule
\textbf{GIN(E)} & 99.3 & 95.1 & 19.7 & 20.0 & 84.3 & 53.3 & 77.6 & 49.5 \\
\textbf{PGN}    & 99.5 & 99.6 & 29.9 & \textbf{26.9} & 97.2 & \textbf{92.0} & 84.6 & \textbf{75.6} \\
\textbf{RecGNN} & 99.5 & 99.3 & 18.7 & 13.5 & 76.0 & 25.0 & 66.6 & 25.7 \\
\midrule
\rowcolor{gray!10} 
\textbf{\fnet}  & \textbf{100.0} & \textbf{100.0} & \textbf{60.6} & 26.4 & \textbf{99.3} & 75.2 & \textbf{97.6} & 38.9 \\ 
\bottomrule
\end{tabular}
\end{table}

\paragraph{Model detail.} \cref{tab:tsp_param_and_time} reports the number of parameters and training time for \fnet on the CLRS benchmark. The model is implemented with \textit{Jax}\citep{jax2018github} based on the official codebase of CLRS. The model is configured with head\_dim = 64, num\_heads = 6, batch\_size=32, and trained for 40K iterations.
\begin{table}[htbp]
\centering
\caption{\textcolor{black}{Computational cost analysis. We report parameter counts, peak GPU memory, and total training time (40K iterations) across different model depths.}}
\label{tab:tsp_param_and_time}
\small
\renewcommand{\arraystretch}{1.2} 
\setlength{\tabcolsep}{8pt}       
\begin{tabular}{c ccc ccc}
\toprule
\multirow{2.5}{*}{\textbf{Depth}} & \multicolumn{2}{c}{\textbf{CLRS Benchmark}} & & \multicolumn{3}{c}{\textbf{TSP Benchmark}} \\
\cmidrule(lr){2-3} \cmidrule(lr){5-7} 
 & \textbf{Params} (M) & \textbf{Time} (min) & & \textbf{Params} (M) & \textbf{GPU Mem} (GB) & \textbf{Time} (min) \\
\midrule
8  & 20.2  & 33.3  & & 17.8  & 8.8  & 19.7  \\
16 & 37.9  & 46.7  & & 35.5  & 12.8 & 30.8  \\
32 & 73.3  & 86.7  & & 70.9  & 23.2 & 53.0  \\
64 & 144.2 & 160.0 & & 141.7 & 46.7 & 100.0 \\
96 & 215.0 & 233.3 & & 212.6 & 67.5 & 146.7 \\
\bottomrule
\end{tabular}
\end{table}

\paragraph{LLMs on CLRS.} \citet{mcleish2024benchmarking} evaluate ChatGPT on CLRS tasks and report solving problems using Python. These results are not directly comparable to ours: generated programs are executed externally, whereas our setting evaluates an end-to-end learned neural processor. Program generation remains part of the former system's problem-solving process; the distinction concerns where the algorithmic computation is carried out.

\subsection{TSP: Pairwise Relational Optimization}
\label{sec:appendix_tsp_details} 
\paragraph{General vs. metric TSP}
A crucial distinction in the Traveling Salesman Problem (TSP) exists between the general (non-metric) TSP and its more structured special case, the metric TSP. The metric TSP, which includes Euclidean TSP, requires the triangle inequality: for any $i,j,k$, the direct edge is no longer than the two-edge route through $j$, $w_{ik}\leq w_{ij}+w_{jk}$. This structure can be exploited algorithmically; in particular, Euclidean TSP admits a polynomial-time approximation scheme (PTAS) \citep{arora1998ptas}.

The \textbf{general TSP}, in contrast, places no such restrictions on edge weights. It is less structured and lacks the approximation guarantees and geometric inductive biases available in metric and Euclidean TSP. Success on arbitrary-weight instances therefore tests performance without reliance on those geometric heuristics. In our work, we provide a comprehensive analysis by testing and comparing our model's performance on both variants, with a particular focus on the general TSP to demonstrate the robustness of our approach.

\paragraph{Solvers for Ground Truth and Baseline}
\begin{itemize}[leftmargin=*,topsep=0pt,itemsep=-2pt]
    \item \textbf{Concorde:} We use Concorde, an exact branch-and-cut solver for symmetric TSP instances \citep{applegate1998solution}, to obtain reference optimal tours for the generated instances.
    \item \textbf{Linkern:} Linkern implements the Chained Lin--Kernighan heuristic \citep{applegate2003chained}, building on the original Lin--Kernighan local-search method \citep{lin1973effective}. We use the configured Linkern run as a non-learned baseline.
\end{itemize}

\paragraph{Data Generation and Filtering Pipeline}
Our data generation process is designed to create challenging and unambiguous test cases. First, we generate instances with the number of nodes $N$ ranging from 10 to 200. We create two types of instances: \textbf{Euclidean TSP}, for which we generate $N$ unique 2D integer coordinates $(X, Y)$ uniformly from $[1, 100]$, and \textbf{general TSP} (also known as explicit TSP), for which we directly generate an $N \times N$ distance matrix with integer weights uniformly sampled from $[1, 100]$. For Euclidean instances, the raw coordinates are directly passed to TSP solvers, so they can leverage the geometric structure of the problem to optimize the solution. But the coordinates are converted into a distance matrix before feeding to our model, to ensure it does not access the geometric information.

Second, we solve each instance with Concorde. For difficult cases, our data-generation wrapper retries unresolved runs using multiple recorded configurations and seeds. This stage used over 1 million CPU-core-hours and yielded 40 million solved instances.

\begin{figure}[ht]
    \centering
    \includegraphics[width=0.75\textwidth]{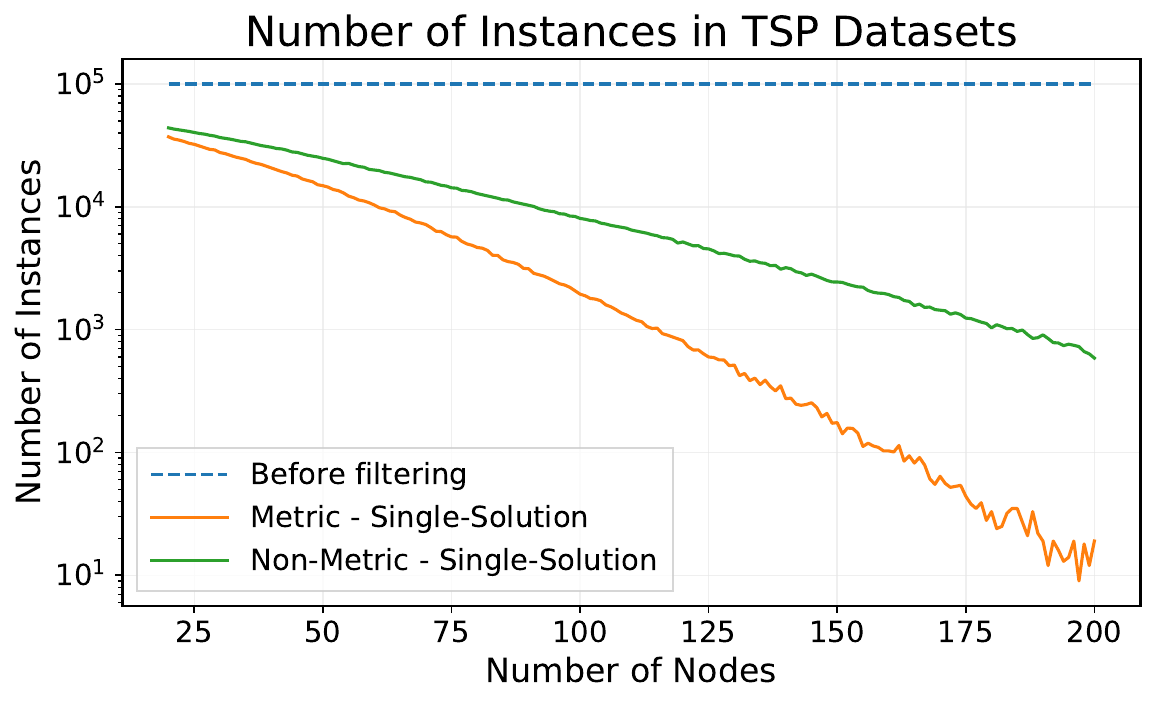}
    \caption{The data distribution of Test Datasets}
    \label{fig:tsp_num_data_points}
\end{figure}

Third, to reduce ambiguity from multiple optimal tours, we perturb each instance slightly and re-solve it, retaining cases for which the same tour is recovered. This procedure reduces solution multiplicity in the evaluated subset but does not certify that the original instance has a unique optimum. About 90\% of instances are removed. We then run Linkern on each retained instance to obtain the comparison baseline. This stage also takes over 1 million CPU-core-hours, yielding 4 million retained instances. \cref{fig:tsp_num_data_points} shows their distribution.

Finally, we partition the dataset into a training/validation set with $N \le 100$ and a test set with $100 < N \le 200$. This split is designed to evaluate the model's ability to generalize to larger problem sizes than seen during training.

\begin{figure}[ht]
    \centering
    \includegraphics[width=0.75\textwidth]{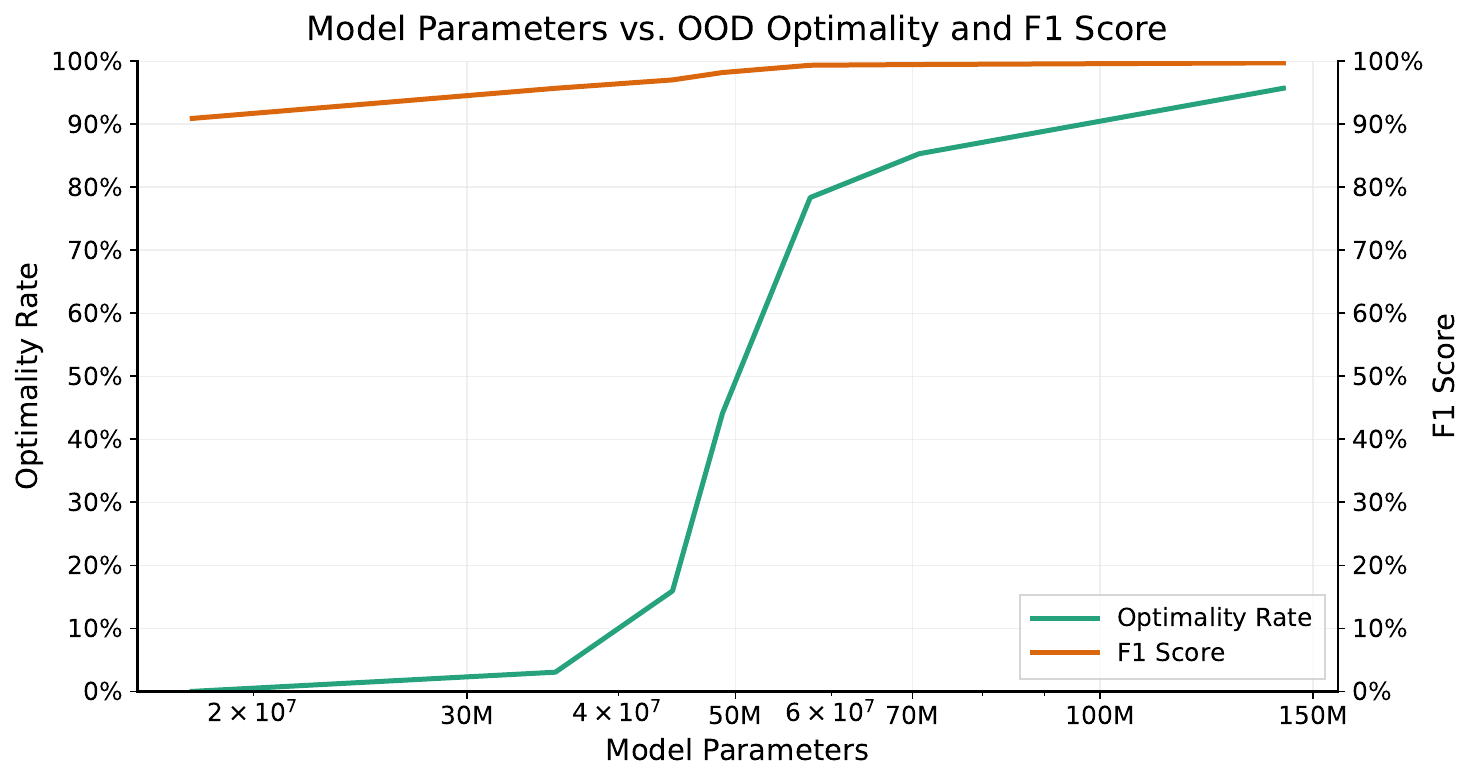}
    \caption{In the evaluated model configurations, optimality rate rises from $0\%$ to $96\%$ while F1 remains above $90\%$, motivating the use of optimality-based metrics in addition to F1.}
    \label{fig:tsp_f1}
\end{figure}

\paragraph{Baseline and Metrics.}
\citet{dwivedi2023benchmarking} introduced a Euclidean-TSP task in the Benchmarking GNNs suite, for which F1 is a reported metric. In our exact-tour experiment, \cref{fig:tsp_f1} shows that F1 can remain above $90\%$ while optimality rate varies substantially; F1 alone therefore does not distinguish these tour-level failure modes. To address this, we adopt the \emph{optimality rate} as the evaluation metric for model predictions, defined as whether the predicted tour is a valid Hamiltonian cycle with a total length equal to the TSP solution produced by the Concorde solver. We also report the detailed optimality gap in \cref{tab:tsp_detailed_appendix}.

\paragraph{Model Details} Despite the relatively high spatiotemporal complexity of \fnet, a well-optimized kernel makes end-to-end training feasible. \cref{tab:tsp_param_and_time} reports the number of parameters, memory consumption, and training time for \fnet on the TSP dataset. The model is configured with head\_dim = 64, num\_heads = 6, batch\_size=1, and bfloat16 precision, and is trained for 40K iterations with a gradient accumulation factor of 8.

\begin{table}[h]
\centering
\caption{Detailed Performance on TSP-100 and TSP-200 benchmarks. We report the gap to the Linkern solver, the percentage of valid tours found (Valid Rate), the optimality gap to the ground truth (Gap), and the average inference time. For multi-solution settings, we evaluate performance with different sampling budgets ($k=1, 5, 10$).}
\label{tab:tsp_detailed_appendix}
\resizebox{\textwidth}{!}{%
\begin{tabular}{llccccccccc}
\toprule
\multirow{2}{*}{\textbf{Setting}} & \multirow{2}{*}{\textbf{Type}} & \multirow{2}{*}{\textbf{Samples}} & \multicolumn{4}{c}{\textbf{TSP-100}} & \multicolumn{4}{c}{\textbf{TSP-200}} \\
\cmidrule(lr){4-7} \cmidrule(lr){8-11}
 &  &  & \textbf{Linkern Gap (\%)} & \textbf{Valid (\%)} & \textbf{Gap (\%)} & \textbf{Time (s)} & \textbf{Linkern Gap (\%)} & \textbf{Valid (\%)} & \textbf{Gap (\%)} & \textbf{Time (s)} \\
\midrule
\multirow{2}{*}{\shortstack{\textbf{Single}\\\textbf{Solution}}} 
 & Metric & 1 & 0.001 & 98.11 & 0.001 & 0.07 & 0.017 & 89.23 & 0.002 & 3.03 \\
 & Non-Metric & 1 & 0.033 & 97.03 & 0.001 & - & 0.375 & 92.63 & 0.000 & - \\
\midrule
\multirow{6}{*}{\shortstack{\textbf{Multi}\\\textbf{Solution}}} 
 & \multirow{3}{*}{Metric} & 1 & 0.002 & 99.99 & 0.025 & 3.13 & 0.019 & 98.94 & 0.075 & 14.83 \\
 &  & 5 & - & 99.99 & 0.002 & - & - & 99.06 & 0.011 & - \\
 &  & 10 & - & 99.99 & 0.001 & - & - & 99.01 & 0.004 & - \\
 \cmidrule{2-11}
 & \multirow{3}{*}{Non-Metric} & 1 & 0.029 & 99.83 & 0.062 & - & 0.290 & 99.67 & 0.060 & - \\
 &  & 5 & - & 99.85 & 0.004 & - & - & 99.69 & 0.003 & - \\
 &  & 10 & - & 99.85 & 0.001 & - & - & 99.71 & 0.001 & - \\
\bottomrule
\end{tabular}%
}
\end{table}

\subsection{Transfer to Molecular and Long-Range Relational Tasks}
\label{sec:appendix_real_world}

\paragraph{ZINC.}
Among the baselines reported in \Cref{tab:exp_real_appendix}, \fnet attains the lowest MAE on both the ZINC-subset and full datasets.

\begin{figure*}[htbp]
\centering
\begin{minipage}[t]{0.32\textwidth}
  \centering
  \includegraphics[width=\linewidth]{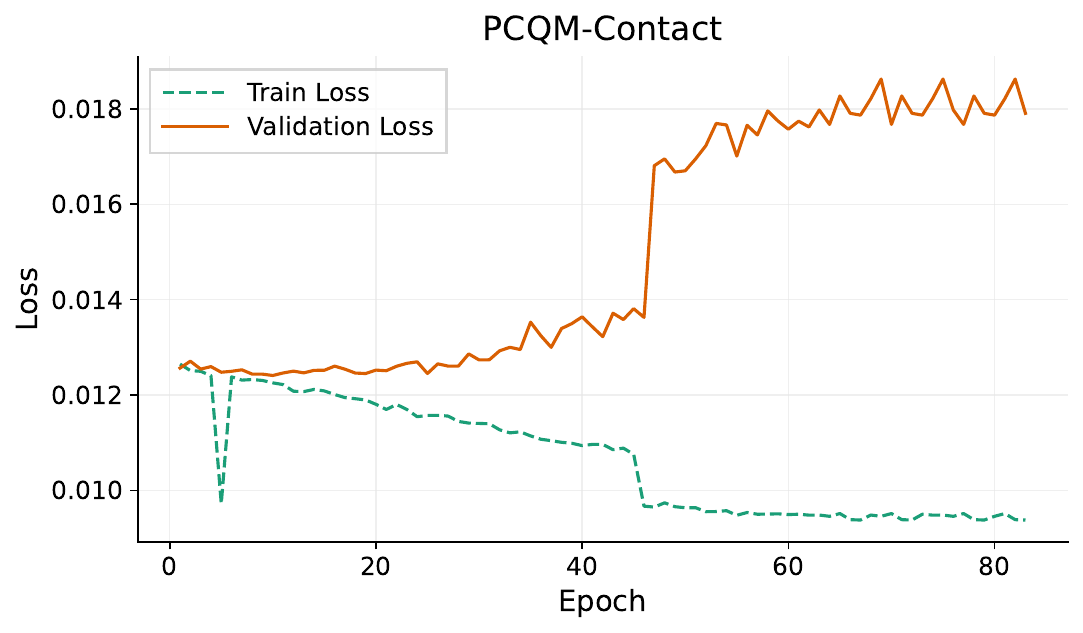}

\end{minipage}
\hfill
\begin{minipage}[t]{0.32\textwidth}
  \centering
  \includegraphics[width=\linewidth]{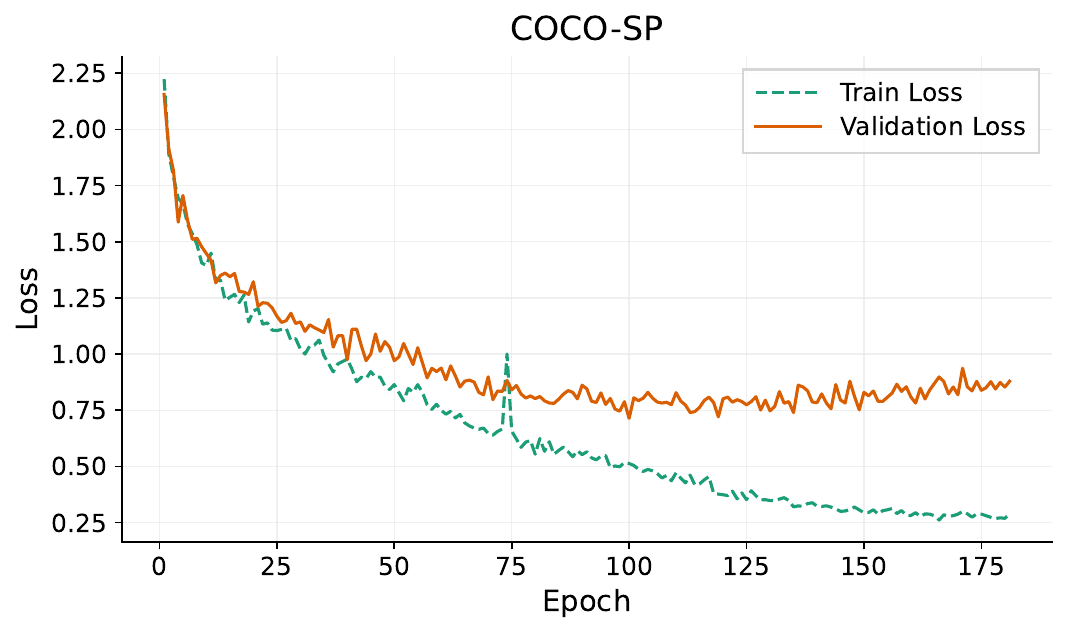}

\end{minipage}
\hfill
\begin{minipage}[t]{0.32\textwidth}
  \centering
  \includegraphics[width=\linewidth]{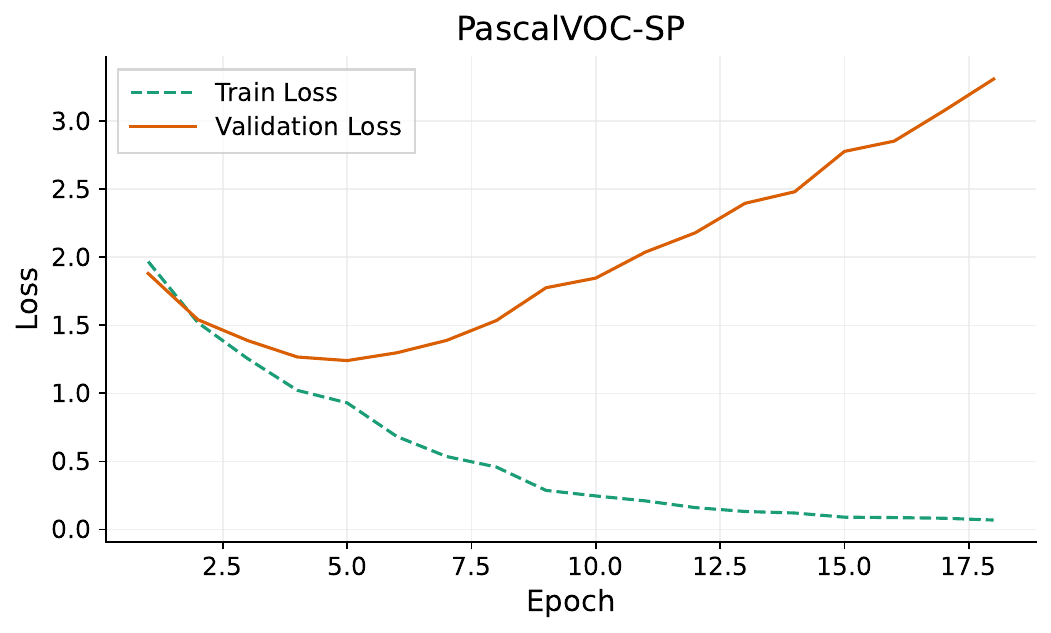}

\end{minipage}

\vspace{0.5em} 

\begin{minipage}{0.66\textwidth} 
  \centering
  \begin{minipage}[t]{0.48\textwidth}
    \centering
    \includegraphics[width=\linewidth]{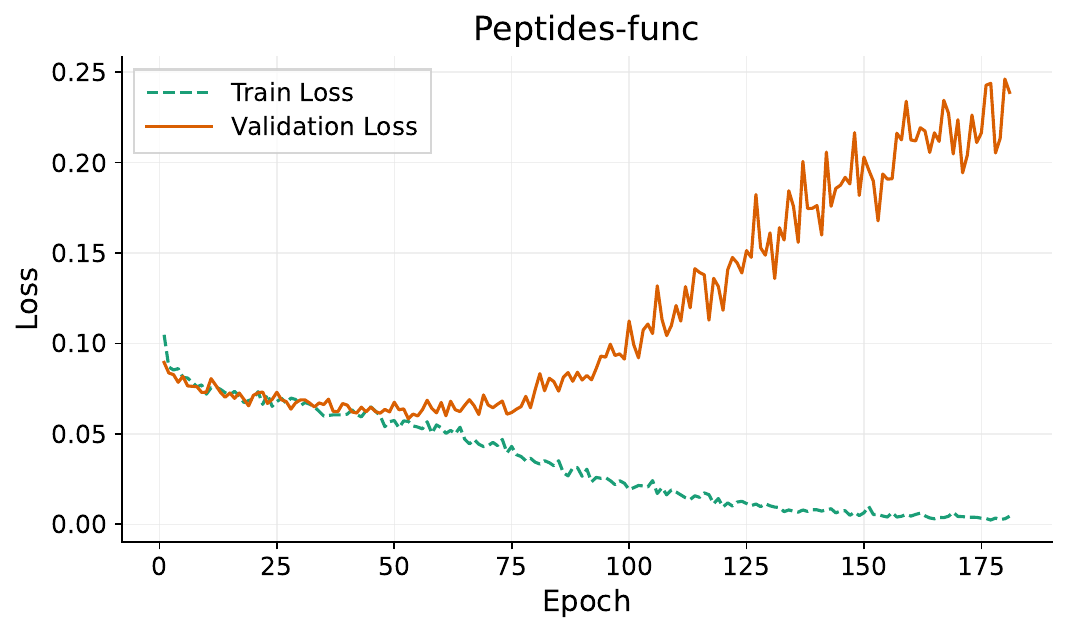}

  \end{minipage}
  \hfill
  \begin{minipage}[t]{0.48\textwidth}
    \centering
    \includegraphics[width=\linewidth]{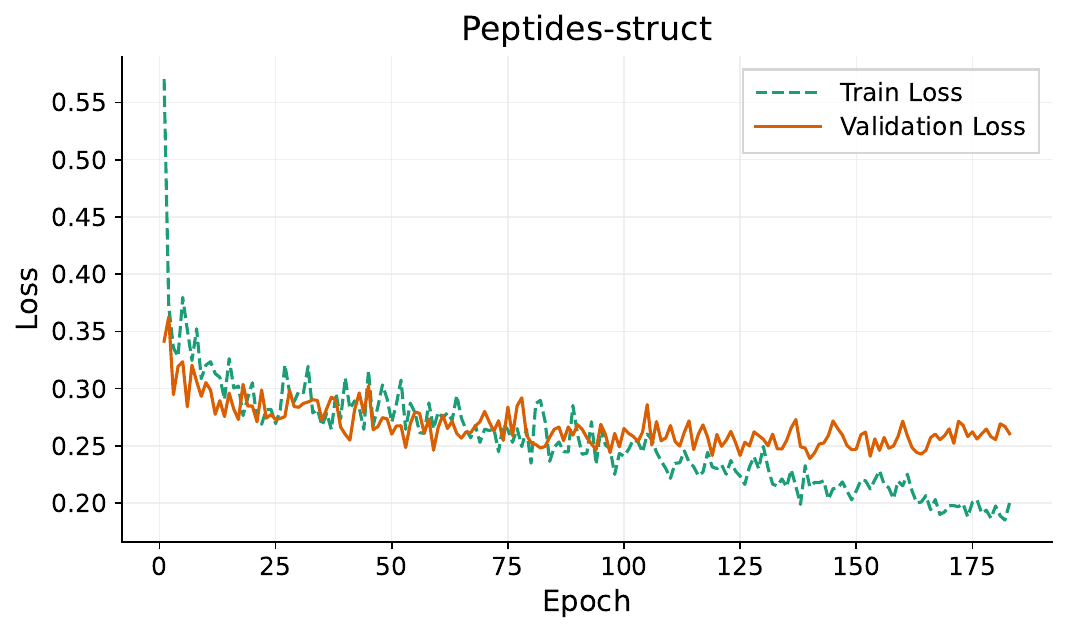}

  \end{minipage}
\end{minipage}

\caption{Training and validation loss on LRGB tasks. The curves demonstrate overfitting: as training epochs increase, training loss consistently declines while validation loss begins to rise. A full search for regularization hyperparameters to mitigate this was limited by computational resources.}
\label{fig:lrgb_train_val_loss}
\end{figure*}

\paragraph{LRGB.}
On the Long-Range Graph Benchmark (LRGB) \citep{dwivedi2022long}, \fnet obtains the strongest reported results in our table on PCQM-Contact and COCO-SP. On the remaining tasks, the displayed training and validation curves (\Cref{fig:lrgb_train_val_loss}) are consistent with possible overfitting, motivating further regularization studies.

\begin{table}[ht]
\centering
\caption{MAE on the ZINC dataset. Lower is better.}
\label{tab:exp_real_appendix}
\begin{tabular}{ll|cc}
\toprule
\multirow{2}{*}{Class} & \multirow{2}{*}{Model} & \multicolumn{2}{c}{ZINC} \\
& & Subset & Full  \\ \midrule
\multirow{3}{*}{GNNs} & MPNN & .138 & .030 \\
& Subgraph GNN & .110 & .028 \\
& Local 2-GNN & \cellcolor{green!20} .069 & \cellcolor{green!20} .024  \\ \midrule

\multirow{2}{*}{\textcolor{black}{GTs}} & \textcolor{black}{Graphormer \citep{ying2021transformers}} & \textcolor{black}{.081} & \textcolor{black}{.025} \\ 
& \textcolor{black}{$\text{ET}_{\text{+RRWP}}$\citep{muller2024towards}} & \textcolor{black}{.059} & \textcolor{black}{.024} \\
\midrule

\multirow{1}{*}{Ours} &  \textbf{\fnet} & \cellcolor{cyan!20} \textbf{.058}& \cellcolor{cyan!20} \textbf{.016}\\
\bottomrule
\end{tabular}
\end{table}

\end{document}